\documentclass{article}

\usepackage{iclr2027_conference,times}
\iclrfinalcopy

\usepackage[T1]{fontenc}
\usepackage[utf8]{inputenc}
\usepackage{url}
\usepackage{hyperref}
\usepackage{xcolor}

\usepackage{amsmath, amssymb, amsfonts, amsthm}
\usepackage{bbm}
\usepackage{booktabs}
\usepackage{multirow}
\usepackage{graphicx}
\usepackage{subcaption}
\usepackage{xspace}
\usepackage{enumitem}
\usepackage{algorithm}
\usepackage{algpseudocode}
\usepackage{xurl}
\usepackage{rotating}
\usepackage[table]{xcolor}
\usepackage{wrapfig}

\newcommand{\sft}{SFT\xspace}
\newcommand{\Sft}{Supervised fine-tuning\xspace}
\newcommand{\lora}{LoRA\xspace}
\newcommand{\ourmethod}{\textsc{GB-SFT}\xspace}
\newcommand{\classbalance}{$\textsc{GB-SFT}_{\text{class}}$\xspace}
\newcommand{\spanbalance}{$\textsc{GB-SFT}_{\text{span}}$\xspace}

\newcommand{\btheta}{\boldsymbol{\theta}}

\newtheorem{lemma}{Lemma}
\newtheorem{corollary}{Corollary}

\newcommand{\hqa}{\textsc{Housing-QA}\xspace}                         
\newcommand{\ncreal}{\textsc{NC-Real}\xspace}                
\newcommand{\realval}{\textsc{RealVal}\xspace}                 
\newcommand{\openfda}{\textsc{OpenFDA}\xspace}             
\newcommand{\tulu}{\textsc{tulu}\xspace}  

\newcommand{\sftgo}{\textsc{\sft-GO}\xspace}                         
\newcommand{\anchoronly}{\textsc{Anchor-Only}\xspace}                
\newcommand{\maskedsft}{\textsc{Masked-SFT}\xspace}                 
\newcommand{\sftkl}{\textsc{\sft + KL}\xspace}             
\newcommand{\dft}{\textsc{DFT}\xspace}    
\newcommand{\asft}{\textsc{ASFT}\xspace}   
\newcommand{\gated}{\textsc{Gated}\xspace}
\newcommand{\gatedinv}{\textsc{Gated-Inv}\xspace}

\newcommand{\qwen}{Qwen3.5-9B-Base\xspace}
\newcommand{\llama}{Llama 3.1-8B\xspace}

\newcommand{\best}[1]{\textbf{#1}}

\title{Group-Balanced Supervised Fine-Tuning for Knowledge-Intensive Domains}

\author{Mark Heimann \\
Ivo AI \\
\texttt{mark.heimann@ivo.ai} \\
\And
Bruno Jacob \\
Ivo AI \\
\texttt{bruno@ivo.ai} \\
}

\begin{document}
\maketitle

\begin{abstract}
Supervised fine-tuning (\sft) weights every supervised token equally, so on knowledge-dense corpora the loss budget follows token count rather than importance: on a legal QA benchmark, citation strings absorb 81\% of the supervised loss weight budget while the answers being taught absorb 5\%. We present Group-Balanced SFT (\ourmethod), which equalizes the loss budget across the \emph{semantic roles} of tokens within a target sequence. With one tunable dial and one structural switch, \ourmethod contains standard \sft, class- and span-level balancing as corners whose geometry we characterize exactly. On benchmarks in law, medicine, code, and mathematics, we show across model families that: (i) \ourmethod injects new knowledge better than uniform \sft, pointwise token weighting, memory editing, and matched-budget data mixing; (ii) it matches the injection of masking out everything but the knowledge payload  while retaining the citation recall that masking destroys (0.98 vs. 0.38); and (iii) role-aware grouping carries the gain: label-free partitions recover at most a third of it, an LLM tagger nearly all where the groups are lexically distinctive; and and (iv) the budget governs what
fine-tuning \emph{breaks} as well as what it learns, acting as a monotone lever on the safety-guardrail erosion \sft causes, at matched task accuracy. We release two new contamination-controlled knowledge-injection benchmarks with validated ground truth.

\end{abstract}

\section{Introduction}
\label{sec:introduction}

Large language models achieve strong general-purpose performance, but deployment in knowledge-intensive domains such as law or medicine requires learning specialized facts \citep{lewis2020rag, singhal2023medpalm}. These domains also fix \emph{how} facts are written: citation grammars, structured product labels and clinical schemas are mandated externally, leaving the fact to be learned as a small span inside a document that is largely required boilerplate. \Sft, the most common adaptation strategy \citep{ouyang2022instructgpt, zhou2023lima}, weights every supervised token equally and spends most of its loss budget on that boilerplate.

\begin{figure}[t]
    \includegraphics[width=\linewidth]{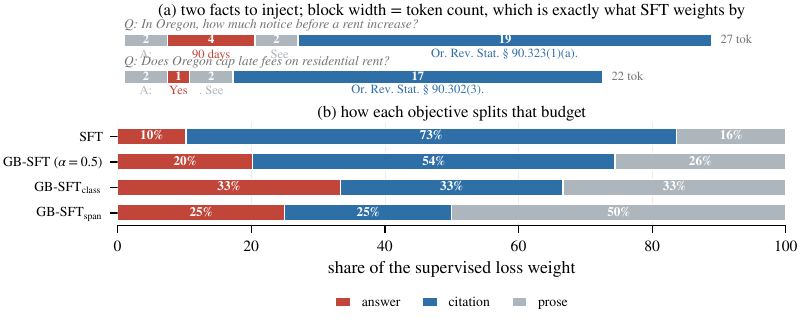}
    \caption{Example of how our grouping budgeting strategies upweight the knowledge payload in a 2-fact corpus. \sft weights the knowledge payload by its share of tokens. \classbalance gives each class (color) an equal budget, and \spanbalance gives each span an equal budget ($\propto$ number of red blocks). Our unified formulation also gives a tunable $\alpha$ knob to gradually increase the class balance.}
    \label{fig:budget-example}
\end{figure}

As a motivating illustration, we consider the task of fine-tuning a model on legal QA~\citep{zheng2025reasoning}, and break down a two-example corpus of legal questions in Fig.~\ref{fig:budget-example}. We call the tokens that state the actual answer the \emph{payload} and the surrounding boilerplate (citations, templated prose) the \emph{scaffold}. By token count, the payload forms a very small portion of the examples compared to scaffold (a numeral-dense citation number, and some extra short prose). This shows that not all tokens in the supervised target carry equal knowledge payload.

A natural solution is to use a fine-tuning objective with some kind of non-uniform token weighting scheme. Some prior works~\citep{lin2024rho, wu2025dft, liu2026eksft, wang2026prift} learn pointwise token importance scores. Our key insight hinges on the fact that tokens in knowledge-dense corpora can be semantically grouped, most readily by the nature of fact construction itself (e.g. as part of the answer, citation, or prose in legal QA). We propose a unified group-balanced supervised fine-tuning (\ourmethod) framework, in which we theoretically and empirically study different grouping strategies and ways to balance the loss among groups. \ourmethod outperforms both pointwise weighting and generic group heuristics~\citep{kim2025sftgo} on knowledge-dense corpora from multiple domains, achieving a favorable balance between preserving existing knowledge and injecting new knowledge.
We make the following contributions:

\begin{itemize}[leftmargin=*, itemsep=0pt]
    \item New group-balanced fine-tuning objectives that give more weight to the knowledge payload.
    \item Improved group construction methods and theoretical characterization of balancing granularity. 
    \item New benchmarks for measuring knowledge injection across multiple domains.
    \item Empirical insight into the tradeoffs between preserving and injecting knowledge. 
\end{itemize}

\section{Related Work}
\label{sec:related}

\noindent \textbf{Token weighting.}
Dynamic Fine-Tuning (DFT) weights each token's cross-entropy by the policy's own detached probability;
Anchored supervised fine-tuning (ASFT) stabilizes DFT
with a uniform token-level Kullback–Leibler (KL) anchor to the frozen base model \citep{zhu2025asft}. Related online-weighting rules include probability thresholding and temperature-scaled variants \citep{lin2026sft, zhang2026towards, diao2026entropy}. 
Rho-1 selects tokens by excess loss against a curated-data reference \citep{lin2024rho}, PriFT weights tokens by prior support under the pretrained checkpoint \citep{wang2026prift}, and EKSFT masks high-entropy tokens from imitation while regularizing them toward the reference \citep{liu2026eksft}. These methods share a pointwise premise (a token's weight should be a function of its own difficulty or plausibility) and can miss group-level structure.

\noindent \textbf{Group-based weighting and class imbalance.} Closest to our approach, \textsc{Sft-GO} partitions tokens into importance groups and optimizes a mixture of worst-group and average loss: a token-level Group-DRO objective \citep{kim2025sftgo, sagawa2020distributionally}.
Rather than adaptively emphasizing the hardest group, a heuristic we find points the wrong way on fact-editing tasks, we balance the loss budget each role receives.
Our budget correction descends from inverse-frequency and class-balanced losses in classification \citep{cui2019class}, and contrasts with difficulty-based schemes such as focal loss \citep{lin2017focal}, which upweight hard tokens (the opposite of what we need). In App.~\ref{app:related} we give an extended discussion of related work, \citet{ye2025disentangling} which uses a weighting from group loss statistics and \citet{hsieh2023distilling} which extracts LLM rationales as additional supervision; we experimentally compare in App.~\ref{app:ye-baseline}.

\noindent \textbf{Span-level supervision units.} Treating spans rather than tokens as the unit of masking or supervision has precedent in pretraining: salient span masking targets entity spans for knowledge acquisition \citep{guu2020realm}, and SpanBERT and PMI-masking mask correlated spans%
~\citep{joshi2020spanbert, levine2021pmi}.
On the aggregation-unit question, recent work shows that the normalization unit of an RL objective silently changes what is optimized \citep{liu2025drgrpo}; our identity for pooled- versus span-normalized CE is the \sft instance of this observation.

\noindent \textbf{Knowledge injection, forgetting, and post-hoc robustness.} Fine-tuning on new facts can promote hallucination \citep{gekhman2024finetuning}, and knowledge becomes extractable only under sufficient paraphrase and format diversity \citep{allenzhu2024physics}; our evaluation accounts for these considerations.
Memory-editing methods modify facts directly in weights \citep{meng2022rome, meng2023mass}, though they do not produce task training. We compare against continual-learning levers such as rehearsal/replay \citep{chaudhry2019tiny, kotha2026replaying} and and post-hoc weight interpolation \citep{wiseft}, which we find preserve existing more than acquire new knowledge. 

\noindent \textbf{Legal NLP benchmarks.} We build on Housing Statute QA \citep{zheng2025reasoning} and evaluate statutory value injection on real statute excerpts, akin to LegalBench \citep{guha2023legalbench}, CaseHOLD \citep{zheng2021casehold}, and SARA \citep{holzenberger2020sara}. We target parametric knowledge, unlike retrieving legal knowledge into context for agents \citep{harveylab}.

\newcommand{\numTokens}{N}
\newcommand{\numClasses}{K}
\newcommand{\numSpans}{P}
\newcommand{\classGrouper}{\mathcal{C}}
\newcommand{\spanGrouper}{\mathcal{S}}
\newcommand{\aclass}{c}
\newcommand{\aspan}{s}
\newcommand{\classes}{\classGrouper}
\newcommand{\spans}{\spanGrouper}
\newcommand{\spansClass}{\spanGrouper_{\aclass}}
\newcommand{\numSpansClass}{P_\aclass}
\newcommand{\numTokensClass}{n_\aclass}
\newcommand{\numTokensSpan}{n_\aspan}
\newcommand{\token}{t}
\newcommand{\tokenWeight}{\omega_\token}
\newcommand{\CE}{\mathrm{CE}}
\newcommand{\tokenloss}{\CE_\token(\btheta)}

\section{Method}
A standard \sft loss weighs each supervised token equally, computing the average negative log likelihood (or cross-entropy loss) of the gold token over all tokens:
\begin{equation}
\label{eq:sft}
    \mathcal{L}_{\sft}(\btheta) = \frac{1}{\numTokens} \sum_\token -\log \pi_{\btheta}(y^*_\token \mid x, y^*_{<\token}) = \frac{1}{\numTokens} \sum_\token \tokenloss,
\end{equation}

\noindent where $\btheta$ denotes the model parameters, $\pi_{\btheta}$ the model's conditional next-token distribution, $x$ the input prompt, $y^*_\token$ the gold token at position $\token$, $y^*_{<\token}$ the gold tokens preceding it, $\numTokens$ the number of supervised tokens, and $\tokenloss$ the resulting per-token cross-entropy loss.

This fails to account for the fact that some groups of tokens may carry more knowledge payload than others. Thus, we will show how to partition tokens into groups and balance the loss accordingly.
\label{sec:method}

\subsection{Token Groupings}
\label{sec:group-formulation}

Grouping is specified by two maps over the supervised tokens of a datum. A \emph{class grouper} $\classGrouper$ assigns each token one of $\numClasses$ classes, and a \emph{span grouper} $\spanGrouper$ partitions the tokens into $\numSpans$ spans. We require that $\spanGrouper$ \emph{refine} $\classGrouper$: every span lies entirely within a single class. Both groupers admit supervised and unsupervised realizations.
Writing $\pi_{\mathrm{ref}}$ for the frozen pretrained
reference model, we consider the following class- and span-level groupers:

\begin{center}\small
\begin{tabular}{@{}llp{0.42\columnwidth}@{}}
\toprule
\textbf{Grouper} & \textbf{Realization} & \textbf{How it is obtained} \\
\midrule
$\classGrouper$ & Category          & fact structure, e.g. \{answer, citation, prose\} \\
              & Citation regex            & a grammar over citation surface form \\
              & LLM tagger                & a frontier LLM asked to label token roles \\
              & Surprisal & $\numClasses$ quantile buckets of $\pi_{\mathrm{ref}}$; label-free \\
\midrule
$\spanGrouper$ & Same class         & maximal contiguous runs of constant class, i.e.\ induced by $\classGrouper$ \\
              & Detector       & cross-fit logistic regression on (surprisal, entropy) in a $\pm 5$ window \\
\bottomrule
\end{tabular}
\end{center}

In our labeled runs $\spanGrouper$ is induced by $\classGrouper$ through
contiguity, which makes refinement automatic. That chaining is a property of
the instantiation and not of the framework: the Detector grouper produces spans
with no class labeling at all, and the Surprisal grouper produces classes
with no span structure.
In Fig.~\ref{fig:budget-example}, $\classGrouper$ assigns each token to $\{$answer, citation, prose$\}$, and $\spanGrouper$ is the maximal runs of constant class (colored blocks). 

\subsection{Group-Balanced Loss Budgeting}
\label{sec:budget-formulation}
Now our $\numTokens$ supervised tokens each belong to a class $\aclass \in \classes$ and span $\aspan \in \spans$, where class $\aclass$ has $\numTokensClass$ tokens and $\aspan$ has $\numTokensSpan$ tokens.
We write $\classes$ and $\spans$ both for the groupers and for the partitions they induce, and identify a class $\aclass$ (resp.\ a span $\aspan$) with its set of token positions, so that $\token \in \aclass$ and $\token \in \aspan$ are well defined.
Our proposed Group-Balanced Supervised Fine-Tuning, \ourmethod, equalizes the loss budget across the coarse class-level groups, or the refined span-level groups. 

\noindent \textbf{Group-Balanced by Class.} To get the loss function for our \classbalance strategy, we compute the mean per-token loss for each class and then take the mean per-class loss across all classes:  
\begin{equation}
\label{eq:balanced}
\mathcal{L}_\text{class-balanced}(\btheta) = \frac{1}{\numClasses} \sum_{\aclass \in \classes} \frac{1}{\numTokensClass} \sum_{\token \in \aclass} \tokenloss.
\end{equation}

\noindent \textbf{Group-Balanced by Span}. To get the loss function for our \spanbalance strategy, we compute the mean per-token loss for each span and then take the mean per-span loss across all spans:
\begin{equation}
\label{eq:spanmean}
\mathcal{L}_\text{span-balanced}(\btheta) = \frac{1}{\numSpans} \sum_{\aspan \in \spans} \frac{1}{\numTokensSpan} \sum_{\token \in \aspan} \tokenloss.
\end{equation}

In Fig.~\ref{fig:budget-example} we visualize a toy housing law fact corpus. Under the tokenization of the \qwen tokenizer~\citep{yang2025qwen3} we show the breakdown of tokens across the categories of \emph{answer} (the main knowledge payload), \emph{citation}, and \emph{prose} (the latter two being scaffolds for the knowledge), and the loss breakdown under ordinary \sft versus our \classbalance and \spanbalance.  This clearly shows the increased emphasis that group balancing places on the knowledge payload.

\paragraph{A unified formulation.} Both balanced loss functions, along with ordinary \sft, can be unified in a single formulation in which we can modulate $\beta$, the pooling
unit within a class, and $\alpha$ the budget split across classes. Define the following terms for a token $\token$ lying in span $\aspan$ and class $\aclass$:

\begin{equation}
u_\aspan = \numTokensSpan^{\beta}, \qquad
m_\aclass = \sum_{\aspan \subset \aclass} u_\aspan, \qquad
M = \sum_{\aclass} m_\aclass^{\alpha},
\qquad
\tokenWeight = \frac{m_\aclass^{\alpha}}{M}\cdot\frac{u_\aspan}{m_\aclass}\cdot\frac{1}{\numTokensSpan} = \frac{1}{M}\, m_\aclass^{\alpha-1}\, \numTokensSpan^{\beta-1},
\label{eq:nested-weight}
\end{equation}

The term $\frac{m_\aclass^{\alpha}}{M}$ controls class budget, $\frac{u_\aspan}{m_\aclass}$ controls span budget within a class, and $\frac{1}{\numTokensSpan}$ is the (uniform) budget given to tokens within a span. 
These combine to yield token weights $\tokenWeight$, where $\sum_\token \tokenWeight = 1$ at every $(\alpha, \beta)$, for our general group-balanced \ourmethod loss formulation:
\begin{equation}
\label{eq:gb-sft}
    \mathcal{L}(\btheta) = \sum_\token \tokenWeight \tokenloss,
\end{equation}

where $\alpha$ interpolates between budgeting by class mass and budgeting by class count: at $\alpha=1$ each class receives budget in proportion to its mass $m_\aclass$ (at $\beta=1$, exactly its token count), while at $\alpha=0$ the mass cancels and every class receives $1/\numClasses$ regardless of size. $\beta$ does the same at the span level: at $\beta=0$ each span counts once, however long. Our familiar objectives are special cases of Eq.~\ref{eq:gb-sft}: $(\alpha, \beta) = (1,1)$ yields \sft, $(0,1)$ yields \classbalance, and $(1,0)$ yields \spanbalance.

The two exponents are not symmetric: $\alpha$ is the dial, $\beta$ a structural choice of which unit carries a budget, exercised only at its endpoints. App.~\ref{app:alpha-beta} develops this and connects the pooling-unit axis to prior work.

While we apply this group-based allocation to the supervised loss, we present a formulation for using group allocation to additionally partition a KL anchor~\citep{zhu2025asft} in App.~\ref{app:anchor-strength}.

\subsection{Analysis}
\label{sec:theory}
\newcommand{\spanCE}[1]{\overline{\CE}_{#1}}
\newcommand{\Covhat}{\widehat{\mathrm{Cov}}}
\newcommand{\CV}{\mathrm{CV}}
\newcommand{\Lpool}{\mathcal{L}_{\mathrm{pool}}}
\newcommand{\Lspan}{\mathcal{L}_{\mathrm{span}}}
\newcommand{\spanloss}{\spanCE{\aspan}(\btheta)}
\newcommand{\spanLength}{\numTokensSpan}
\newcommand{\avgSpanLength}{\bar{n}_{\aclass}}
\newcommand{\avgSpanLoss}{\mu_{\aclass}}
\newcommand{\sdSpanLength}{\sigma_{n}}
\newcommand{\CVspan}{\CV(n)}
\newcommand{\classBudget}{p_{\aclass}}

\paragraph{How the loss moves in $(\alpha, \beta)$.}

Let $\spansClass$ denote the spans of class $\aclass$. Group the weights of Eq.~\ref{eq:nested-weight} into a class budget, a within-class span
distribution, and a span-mean loss:
\begin{equation}
p_\aclass = \frac{m_\aclass^{\alpha}}{M},
\qquad
q_\aclass(\aspan) = \frac{u_\aspan}{m_\aclass},
\qquad
\spanCE{\aspan} = \frac{1}{\numTokensSpan}\sum_{\token \in \aspan} \tokenloss,
\qquad
A_\aclass = \sum_{\aspan \in \spansClass} q_\aclass(\aspan)\, \spanCE{\aspan},
\label{eq:budget-shares}
\end{equation}
so that Eq.~\ref{eq:gb-sft} reads $\mathcal{L} = \sum_\aclass p_\aclass A_\aclass$: $p$ is the share of
the budget class $\aclass$ receives, $q_\aclass$ splits that share among $\aclass$'s spans, and
$A_\aclass$ is $\aclass$'s $\beta$-weighted mean loss, (the token mean over the class at
$\beta = 1$, the mean of its span means at $\beta = 0$). Holding $\btheta$ fixed, the
\ourmethod family admits an exact characterization of how the loss moves in $(\alpha, \beta)$ and
between its corners (App.~\ref{app:derivation}):

\begin{equation}
\frac{\partial \mathcal{L}}{\partial \alpha}
  = \mathrm{Cov}_{p}\big(\log m_\aclass,\; A_\aclass\big),
  \qquad 
  \frac{\partial \mathcal{L}}{\partial \beta}
  = \underbrace{\alpha\,\mathrm{Cov}_{p}\big(\mathbb{E}_{q_\aclass}[\log \numTokensSpan],\; A_\aclass\big)}_{\text{between classes}}
   + \underbrace{\mathbb{E}_{p}\big[\mathrm{Cov}_{q_\aclass}\big(\log \numTokensSpan,\; \spanCE{\aspan}\big)\big]}_{\text{within classes}}.
\label{eq:derivative}
\end{equation}

We thus see that $\alpha$ acts only between classes, $\beta$ acts at both levels, and $\beta$'s between-class term carries the factor $\alpha$, so it vanishes at $\alpha=0$. 

The finite counterpart of the within-class term compares the two pooling units at a fixed class budget $\classBudget$, or equivalently at a fixed $p_\aclass$ in Eq.~\ref{eq:budget-shares}.
Within a class, pooled and span-balanced training differ only in how they average the span-mean losses $\spanCE{\aspan}$: pooling weights each span by its length $\spanLength$; span balancing weights spans equally. The gap between them is exactly the length/difficulty covariance:

\begin{lemma}[Covariance identity]
\label{lem:cov-identity}
Fix a class $\aclass$ with spans $\spansClass$, $\numSpansClass = |\spansClass|$, span lengths $\spanLength$, and mean span length $\avgSpanLength = \numTokensClass / \numSpansClass$. Let $\Covhat_{\aspan \in \spansClass}(a, b) = \frac{1}{\numSpansClass}\sum_{\aspan \in \spansClass}(a_\aspan - \bar a)(b_\aspan - \bar b)$ be the unweighted empirical covariance over the spans of $\aclass$. Then, for every $\btheta$,
\begin{equation}
\underbrace{\frac{\sum_{\aspan \in \spansClass} \spanLength\, \spanloss}
{\sum_{\aspan \in \spansClass} \spanLength}}_{\text{token-weighted (pooled)}}
 \;-\;
\underbrace{\frac{1}{\numSpansClass}\sum_{\aspan \in \spansClass} \spanloss}_{\text{span-balanced}}
 \;=\;
\frac{1}{\avgSpanLength}\,
\Covhat_{\aspan \in \spansClass}\!\bigl(\spanLength,\ \spanloss\bigr),
\label{eq:cov-identity}
\end{equation}
 \end{lemma}

This identity is assumption-free: it requires no convexity, smoothness, or boundedness, and holds pointwise at every checkpoint (proof and a gradient-level analogue in App.~\ref{app:proofs}).
A Cauchy-Schwarz argument bounds the difference in losses for a fixed class budget by $\CVspan$, the within-class coefficient of variation of span lengths (Cor.~\ref{cor:cv-bound}, App.~\ref{app:proofs}), explaining why \classbalance and \spanbalance rarely separate on our corpora: small $\CVspan$ forces the allocations close. The sign of the covariance is also informative: it is negative (short spans are harder per token) on our \hqa benchmark~\citep{zheng2025reasoning}, so span balancing shifts budget toward exactly the spans that token-proportional pooling starves.

\section{Experiments}
\paragraph{Base LLMs \& Configuration.} Our main base LLMs are \qwen~\citep{yang2025qwen3} and \llama~\citep{dubey2024llama}.
All arms train low-rank adapters (LoRA) \cite{hu2022lora} ($r{=}128$, $\alpha{=}256$, no dropout) and MLP projections for $3$ epochs at learning rate $2\times10^{-5}$ with token-budgeted micro-batches 
on a single NVIDIA B200 GPU. Our findings hold across learning rates (App.~\ref{app:lr-matched}), LoRA ranks (App.~\ref{app:lora-rank}), and model sizes; \ourmethod helps larger models \emph{more} (App.~\ref{app:model-size}). For fine-tuning results, $\pm$ is one standard deviation over three seeds; frozen base model results are a single deterministic evaluation. 

\paragraph{Baselines.} We compare our class and span-balanced methods, \classbalance and \spanbalance, against three families of published SFT methods, chosen so that each isolates a different axis of the design space rather than merely providing a number to beat. In App.~\ref{app:forgetting} and ~\ref{app:memory-edit} we also consider mechanisms that go beyond token-level loss re-weighing as a further control. 

\noindent \emph{Loss Shaping.} We compare against standard supervised fine-tuning, dynamic fine-tuning \dft ~\citep{wu2025dft}, and anchored supervised fine-tuning \asft ~\citep{zhu2025asft}. 

\noindent \emph{Reference-weighted token selection}. We compare against \textsc{PriFT-Mass} and \textsc{PriFT-Prob} ~\citep{wang2026prift}, \textsc{Eksft}~\citep{liu2026eksft}, and \textsc{Rho-1}~\citep{lin2024rho}.

\noindent \emph{Group-based aggregation}. We compare against \sftgo~\citep{kim2025sftgo}, most in the spirit of our work; in addition to running \sftgo as published, we also consider giving it our default category-based token grouping, denoted as $\sftgo_{\text{oracle}}$.%

We also propose several variants that, inspired by ~\cite{zhu2025asft}, use a KL-based anchoring mechanism to preserve knowledge (App.~\ref{app:anchor-strength}). \maskedsft masks the cross-entropy loss from the largest scaffold classes, \anchoronly applies only the anchor term to the scaffold class, and \sftkl adds a KL anchor term to all classes. \gated applies the anchor with a soft surprisal-weighted gate, and we also test its inversion control \gatedinv. 

\subsection{Benchmark Results}
\label{sec:results}

\paragraph{\hqa: Learning real knowledge with no general prior.}
\begin{table}[t]
\centering\scriptsize
\setlength{\tabcolsep}{1.5pt}
\begin{tabular}{@{}lrrrrrrrr@{}}
\toprule
\textbf{Method} & \multicolumn{4}{c}{\textbf{\qwen}} & \multicolumn{4}{c}{\textbf{\llama}} \\
\cmidrule(lr){2-5}\cmidrule(lr){6-9}
 & AUC-in & AUC-out & cite-in & cite-out & AUC-in & AUC-out & cite-in & cite-out \\
\midrule
\textit{untuned base} & .560 & .541 & .580 & .752 & .507 & .556 & .800 & .816 \\
\midrule
\multicolumn{9}{@{}l}{\textit{group balancing (ours)}} \\
\textbf{\classbalance} & .760\,\tiny{$\pm$.013} & .603\,\tiny{$\pm$.002} & .983\,\tiny{$\pm$.002} & .600\,\tiny{$\pm$.017} & .748\,\tiny{$\pm$.009} & .611\,\tiny{$\pm$.011} & .987\,\tiny{$\pm$.005} & .669\,\tiny{$\pm$.010} \\
\textbf{\spanbalance} & \cellcolor{gray!30}.769\,\tiny{$\pm$.009} & .603\,\tiny{$\pm$.010} & .972\,\tiny{$\pm$.003} & .624\,\tiny{$\pm$.006} & .765\,\tiny{$\pm$.006} & .629\,\tiny{$\pm$.013} & \cellcolor{gray!30}.993\,\tiny{$\pm$.002} & .723\,\tiny{$\pm$.007} \\
\midrule
\multicolumn{9}{@{}l}{\textit{anchored variants (ours)}} \\
\anchoronly & .746\,\tiny{$\pm$.004} & .576\,\tiny{$\pm$.009} & .581\,\tiny{$\pm$.016} & \cellcolor{gray!30}.739\,\tiny{$\pm$.019} & .752\,\tiny{$\pm$.011} & .597\,\tiny{$\pm$.008} & .681\,\tiny{$\pm$.008} & .744\,\tiny{$\pm$.026} \\
\gated & .671\,\tiny{$\pm$.011} & .579\,\tiny{$\pm$.020} & .980\,\tiny{$\pm$.006} & .603\,\tiny{$\pm$.023} & .591\,\tiny{$\pm$.023} & .621\,\tiny{$\pm$.004} & .981\,\tiny{$\pm$.004} & .729\,\tiny{$\pm$.013} \\
\gatedinv & .684\,\tiny{$\pm$.009} & .583\,\tiny{$\pm$.013} & .866\,\tiny{$\pm$.022} & .576\,\tiny{$\pm$.016} & .663\,\tiny{$\pm$.013} & \cellcolor{gray!30}.631\,\tiny{$\pm$.023} & .965\,\tiny{$\pm$.002} & .731\,\tiny{$\pm$.015} \\
\midrule
\multicolumn{9}{@{}l}{\textit{group-based aggregation}} \\
\sftgo & .630\,\tiny{$\pm$.009} & .568\,\tiny{$\pm$.003} & .980\,\tiny{$\pm$.000} & .704\,\tiny{$\pm$.012} & .529\,\tiny{$\pm$.006} & .566\,\tiny{$\pm$.012} & \cellcolor{gray!30}.993\,\tiny{$\pm$.002} & .795\,\tiny{$\pm$.022} \\
\sftgo$_{\text{oracle}}$ & .756\,\tiny{$\pm$.014} & .605\,\tiny{$\pm$.015} & .968\,\tiny{$\pm$.004} & .720\,\tiny{$\pm$.016} & \cellcolor{gray!30}.773\,\tiny{$\pm$.010} & .624\,\tiny{$\pm$.018} & .987\,\tiny{$\pm$.002} & \cellcolor{gray!30}.815\,\tiny{$\pm$.013} \\
\midrule
\multicolumn{9}{@{}l}{\textit{reference-weighted selection}} \\
\textsc{PriFT-mass} & .601\,\tiny{$\pm$.012} & .596\,\tiny{$\pm$.011} & .603\,\tiny{$\pm$.009} & .544\,\tiny{$\pm$.012} & .595\,\tiny{$\pm$.008} & .606\,\tiny{$\pm$.005} & .725\,\tiny{$\pm$.011} & .704\,\tiny{$\pm$.037} \\
\textsc{PriFT-prob} & .644\,\tiny{$\pm$.007} & .582\,\tiny{$\pm$.004} & .845\,\tiny{$\pm$.005} & .603\,\tiny{$\pm$.020} & .670\,\tiny{$\pm$.005} & .622\,\tiny{$\pm$.002} & .881\,\tiny{$\pm$.002} & .731\,\tiny{$\pm$.008} \\
\textsc{Rho-1} & .595\,\tiny{$\pm$.004} & .581\,\tiny{$\pm$.012} & .948\,\tiny{$\pm$.012} & .684\,\tiny{$\pm$.017} & .675\,\tiny{$\pm$.004} & .625\,\tiny{$\pm$.008} & .968\,\tiny{$\pm$.003} & .771\,\tiny{$\pm$.018} \\
\textsc{Eksft} & .693\,\tiny{$\pm$.009} & \cellcolor{gray!30}.607\,\tiny{$\pm$.014} & .665\,\tiny{$\pm$.030} & .396\,\tiny{$\pm$.023} & .670\,\tiny{$\pm$.017} & .597\,\tiny{$\pm$.021} & .863\,\tiny{$\pm$.013} & .700\,\tiny{$\pm$.031} \\
\midrule
\multicolumn{9}{@{}l}{\textit{loss shaping}} \\
\sft & .652\,\tiny{$\pm$.011} & .576\,\tiny{$\pm$.021} & \cellcolor{gray!30}.988\,\tiny{$\pm$.003} & .583\,\tiny{$\pm$.016} & .630\,\tiny{$\pm$.008} & .595\,\tiny{$\pm$.015} & .991\,\tiny{$\pm$.002} & .712\,\tiny{$\pm$.015} \\
\sftkl & .652\,\tiny{$\pm$.008} & .566\,\tiny{$\pm$.004} & .980\,\tiny{$\pm$.003} & .631\,\tiny{$\pm$.007} & .631\,\tiny{$\pm$.004} & .595\,\tiny{$\pm$.019} & .991\,\tiny{$\pm$.005} & .665\,\tiny{$\pm$.025} \\
\dft & .529\,\tiny{$\pm$.035} & .538\,\tiny{$\pm$.028} & .483\,\tiny{$\pm$.037} & .523\,\tiny{$\pm$.041} & .515\,\tiny{$\pm$.016} & .518\,\tiny{$\pm$.032} & .541\,\tiny{$\pm$.024} & .659\,\tiny{$\pm$.022} \\
\asft & .509\,\tiny{$\pm$.010} & .493\,\tiny{$\pm$.008} & .645\,\tiny{$\pm$.039} & .596\,\tiny{$\pm$.062} & .486\,\tiny{$\pm$.005} & .567\,\tiny{$\pm$.016} & .788\,\tiny{$\pm$.021} & .725\,\tiny{$\pm$.007} \\
\midrule
\multicolumn{9}{@{}l}{\textit{control}} \\
\maskedsft & .768\,\tiny{$\pm$.007} & .578\,\tiny{$\pm$.012} & .379\,\tiny{$\pm$.005} & .355\,\tiny{$\pm$.010} & .727\,\tiny{$\pm$.012} & .608\,\tiny{$\pm$.004} & .577\,\tiny{$\pm$.007} & .580\,\tiny{$\pm$.013} \\
\bottomrule
\end{tabular}
\caption{
\textbf{\hqa}. AUC-in is mean within-group AUC
over the 99 held-in question groups containing both labels; AUC-out the same over 82 held-out groups. \emph{cite-in} is top-1 citation recall against nine real distractors on the held-in jurisdictions; \emph{cite-out} the same on held-out. 
\emph{\ourmethod achieves the most knowledge payload injection (AUC-in) on
\qwen; our groupings also help \sftgo$_{\text{oracle}}$ preserve the most
scaffold knowledge (cite-out).}
App.~\ref{app:sftgo-grid}
gives the full comparison across corpora and models. Note also that \maskedsft
comes within $.002$ of \spanbalance's AUC-in on \qwen at the cost of its citation recall
collapsing on both splits and models.%
}
\label{tab:hqa-both}
\end{table}

We build on the Housing Statute QA dataset from~\cite{zheng2025reasoning}, where the goal is to teach a model jurisdiction-specific verifiable facts about housing law. We construct a dataset of 1,176 facts about which \qwen has no confident belief about the correct answer. By construction, each fact decomposes into three classes among which we can budget learning: the answer to a housing law question (Yes/No), a citation, and templated prose. Spans are sequences of contiguous tokens belonging to the same class. Full details of this benchmark are listed in Appendix~\ref{app:housingqa-overview}.

In Tab.~\ref{tab:hqa-both} we compare methods along two axes: knowledge payload injection (measured by AUC within question group) and scaffold knowledge preservation (citation recall). We see the following: 

\noindent \emph{Our methods outperform existing token reweighting baselines in knowledge injection}. \ourmethod outperforms the pointwise baselines and default \sftgo, while being competitive with oracle-grouped \sftgo. Meanwhile, \dft~\citep{wu2025dft} and its variant \asft~\citep{zhu2025asft} sit below the base AUC; they reweigh learning based on what the current policy already considers unsurprising or the reference model already knows, the opposite of what a model needs to learn new knowledge. Meanwhile, \spanbalance gains $+0.209$ AUC-in over base. This is a $3.4\times$ ratio with its gains in $+0.062$ AUC-out, indicating that what was learned is jurisdiction-specific knowledge injection rather than general adaptation. These improvements persist under held-out phrasing (App.~\ref{app:phrasing}).

\noindent \emph{There is a tradeoff between preserving knowledge and injecting new knowledge}. On Qwen, \maskedsft, the highest-performing baseline on held-in AUC, achieves held-in citation recall of 0.379, below an untuned base (0.580) and far below our methods at 0.972 and 0.983. For held-out citation recall, no method compared here exceeds base preservation levels ($0.752$) while also injecting on Qwen, On Llama, our grouping does help \textsc{Sft-GO} maintain them.
\anchoronly comes closest on Qwen, giving up $0.024$ AUC compared to \spanbalance to hold held-out citation recall within $0.013$ of base, by not supervising the scaffold at all (thus not improving the base on held-in citation recall). More details of the anchoring tradeoff are in App.~\ref{app:anchor-strength}.

\paragraph{\ncreal: injecting knowledge absent from pretraining.}
\begin{table}[t]
\centering\scriptsize
\setlength{\tabcolsep}{1.5pt}
\begin{tabular}{@{}lrrrrrrrr@{}}
\toprule
\textbf{Method} & \multicolumn{4}{c}{\textbf{\qwen}} & \multicolumn{4}{c}{\textbf{\llama}} \\
\cmidrule(lr){2-5}\cmidrule(lr){6-9}
 & ovwrite & top-1 & \shortstack[r]{$\log p$\\true} & \shortstack[r]{$\log p$\\taught} & ovwrite & top-1 & \shortstack[r]{$\log p$\\true} & \shortstack[r]{$\log p$\\taught} \\
\midrule
\textit{untuned base} & .133 & .011 & $-2.65$ & $-4.99$ & .094 & .009 & $-2.69$ & --- \\
\midrule
\multicolumn{9}{@{}l}{\textit{group balancing (ours)}} \\
\textbf{\classbalance} & .954\,\tiny{$\pm$.000} & .861\,\tiny{$\pm$.006} & $-$4.36\,\tiny{$\pm$.08} & \cellcolor{gray!30}$-$.43\,\tiny{$\pm$.00} & \cellcolor{gray!30}.947\,\tiny{$\pm$.004} & \cellcolor{gray!30}.892\,\tiny{$\pm$.017} & $-$4.97\,\tiny{$\pm$.09} & $-$.46\,\tiny{$\pm$.01} \\
\textbf{\spanbalance} & .954\,\tiny{$\pm$.003} & .861\,\tiny{$\pm$.007} & $-$4.52\,\tiny{$\pm$.05} & \cellcolor{gray!30}$-$.43\,\tiny{$\pm$.01} & .946\,\tiny{$\pm$.004} & .882\,\tiny{$\pm$.004} & $-$5.03\,\tiny{$\pm$.05} & \cellcolor{gray!30}$-$.44\,\tiny{$\pm$.01} \\
\midrule
\multicolumn{9}{@{}l}{\textit{anchored variants (ours)}} \\
\anchoronly & \cellcolor{gray!30}.963\,\tiny{$\pm$.002} & \cellcolor{gray!30}.877\,\tiny{$\pm$.006} & $-$4.84\,\tiny{$\pm$.12} & $-$.46\,\tiny{$\pm$.01} & .919\,\tiny{$\pm$.006} & .865\,\tiny{$\pm$.009} & $-$4.94\,\tiny{$\pm$.09} & $-$.62\,\tiny{$\pm$.02} \\
\gated & .929\,\tiny{$\pm$.007} & .823\,\tiny{$\pm$.007} & $-$4.48\,\tiny{$\pm$.01} & $-$.60\,\tiny{$\pm$.02} & .906\,\tiny{$\pm$.004} & .845\,\tiny{$\pm$.010} & $-$5.34\,\tiny{$\pm$.14} & $-$.68\,\tiny{$\pm$.03} \\
\gatedinv & .425\,\tiny{$\pm$.010} & .143\,\tiny{$\pm$.006} & $-$3.61\,\tiny{$\pm$.01} & $-$4.50\,\tiny{$\pm$.05} & .300\,\tiny{$\pm$.003} & .119\,\tiny{$\pm$.004} & $-$3.70\,\tiny{$\pm$.01} & $-$5.58\,\tiny{$\pm$.06} \\
\midrule
\multicolumn{9}{@{}l}{\textit{group-based aggregation}} \\
\sftgo & .841\,\tiny{$\pm$.004} & .683\,\tiny{$\pm$.012} & $-$3.28\,\tiny{$\pm$.03} & $-$.94\,\tiny{$\pm$.01} & .852\,\tiny{$\pm$.003} & .733\,\tiny{$\pm$.008} & $-$3.77\,\tiny{$\pm$.08} & $-$.96\,\tiny{$\pm$.00} \\
\sftgo$_{\text{oracle}}$ & .910\,\tiny{$\pm$.016} & .749\,\tiny{$\pm$.010} & $-$3.54\,\tiny{$\pm$.04} & $-$.67\,\tiny{$\pm$.05} & .934\,\tiny{$\pm$.009} & .804\,\tiny{$\pm$.005} & $-$3.88\,\tiny{$\pm$.05} & $-$.62\,\tiny{$\pm$.04} \\
\midrule
\multicolumn{9}{@{}l}{\textit{reference-weighted selection}} \\
\textsc{PriFT-mass} & .125\,\tiny{$\pm$.003} & .021\,\tiny{$\pm$.000} & $-$5.60\,\tiny{$\pm$.04} & $-$10.81\,\tiny{$\pm$.06} & .083\,\tiny{$\pm$.001} & .017\,\tiny{$\pm$.003} & $-$5.55\,\tiny{$\pm$.10} & $-$12.49\,\tiny{$\pm$.28} \\
\textsc{PriFT-prob} & .242\,\tiny{$\pm$.006} & .051\,\tiny{$\pm$.006} & $-$3.90\,\tiny{$\pm$.03} & $-$6.50\,\tiny{$\pm$.02} & .174\,\tiny{$\pm$.003} & .040\,\tiny{$\pm$.003} & $-$3.89\,\tiny{$\pm$.03} & $-$7.70\,\tiny{$\pm$.03} \\
\textsc{Rho-1} & .461\,\tiny{$\pm$.007} & .104\,\tiny{$\pm$.013} & \cellcolor{gray!30}$-$2.87\,\tiny{$\pm$.01} & $-$3.26\,\tiny{$\pm$.04} & .412\,\tiny{$\pm$.004} & .085\,\tiny{$\pm$.005} & \cellcolor{gray!30}$-$2.98\,\tiny{$\pm$.01} & $-$3.71\,\tiny{$\pm$.07} \\
\textsc{Eksft} & .596\,\tiny{$\pm$.013} & .330\,\tiny{$\pm$.015} & $-$2.98\,\tiny{$\pm$.03} & $-$2.11\,\tiny{$\pm$.07} & .616\,\tiny{$\pm$.011} & .338\,\tiny{$\pm$.020} & $-$3.15\,\tiny{$\pm$.04} & $-$2.31\,\tiny{$\pm$.09} \\
\midrule
\multicolumn{9}{@{}l}{\textit{loss shaping}} \\
\sft & .909\,\tiny{$\pm$.003} & .821\,\tiny{$\pm$.005} & $-$4.04\,\tiny{$\pm$.02} & $-$.64\,\tiny{$\pm$.01} & .913\,\tiny{$\pm$.001} & .862\,\tiny{$\pm$.004} & $-$5.01\,\tiny{$\pm$.02} & $-$.61\,\tiny{$\pm$.01} \\
\sftkl & .896\,\tiny{$\pm$.002} & .800\,\tiny{$\pm$.006} & $-$3.67\,\tiny{$\pm$.06} & $-$.68\,\tiny{$\pm$.02} & .899\,\tiny{$\pm$.003} & .843\,\tiny{$\pm$.003} & $-$4.61\,\tiny{$\pm$.03} & $-$.66\,\tiny{$\pm$.00} \\
\dft & .239\,\tiny{$\pm$.011} & .047\,\tiny{$\pm$.006} & $-$7.84\,\tiny{$\pm$.36} & $-$12.56\,\tiny{$\pm$.42} & .109\,\tiny{$\pm$.021} & .027\,\tiny{$\pm$.005} & $-$8.63\,\tiny{$\pm$.16} & $-$17.40\,\tiny{$\pm$.59} \\
\asft & .249\,\tiny{$\pm$.010} & .044\,\tiny{$\pm$.003} & $-$4.72\,\tiny{$\pm$.04} & $-$7.73\,\tiny{$\pm$.18} & .189\,\tiny{$\pm$.016} & .042\,\tiny{$\pm$.006} & $-$4.90\,\tiny{$\pm$.06} & $-$9.01\,\tiny{$\pm$.10} \\
\bottomrule
\end{tabular}
\caption{\textbf{\ncreal}, held-in facts.
\emph{Overwrite} is the fraction of items where the taught (false) value outranks the true one; \emph{top-1} additionally requires it to beat all six distractors. $\log p$ values are mean per-token log-probabilities of each candidate continuing the statute's own prefix, so $\log p_{\text{true}}$ measures how much of the real statutory value survives injection. \emph{\ourmethod injects more knowledge than all published baselines and is competitive or superior to our anchoring variant.}
}
\label{tab:ncreal-both}
\end{table}

Our next benchmark is a controlled study of acquiring knowledge that cannot be in a model's training corpus. We take real statutory prose from the LSC Eviction Laws Database and change the values,
to construct a training dataset of 229 distinct facts from which we form 916 training examples by taking the same altered statute body with four different headers. The evaluation dataset consists of 758 items. Each fact decomposes into two classes: values or prose. Spans are once again sequences of contiguous tokens belonging to the same class. More details of this benchmark, called \ncreal, are in App.~\ref{app:ncreal-overview}.

Our main results are shown in Tab.~\ref{tab:ncreal-both}. On Qwen, we observe particularly good performance by \anchoronly, which achieves the top overwrite and top-1 performance by a very narrow margin over \ourmethod; however, \ourmethod variants outperform it by a considerably wider margin on Llama. Moreover, our methods achieve the top performance compared to all other published baselines on both models. \dft and \asft collapse across both models, again verifying our intuition.%

\paragraph{\realval: injecting real missing knowledge.}
\begin{figure}
\centering
\includegraphics[width=0.8\textwidth]{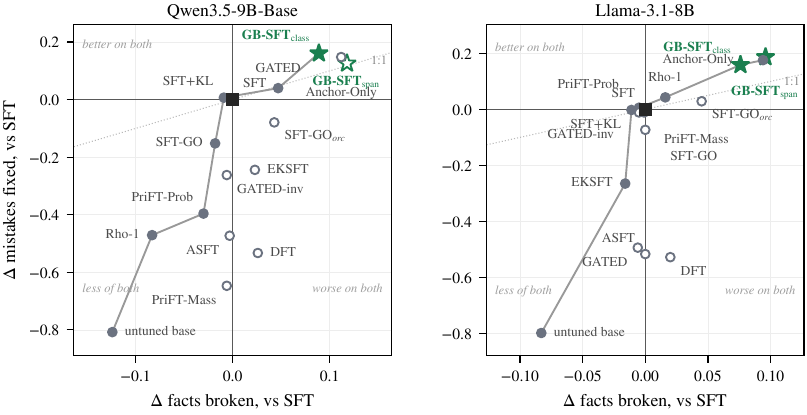}
\caption{\textbf{The exchange rate between learning and breaking knowledge.}
Horizontal axis is $\Delta$ facts \emph{broken} relative to \sft; vertical is
$\Delta$ mistakes \emph{fixed}. \textsc{Sft} is the origin, so an arm's exchange rate
$\Delta\text{correction}/\Delta\text{damage}$ is simply the \emph{slope} of the line
joining it to the origin, and the dotted $1{:}1$ line is the break-even locus: above
it an arm fixes more than it breaks relative to \sft. The solid line is the
Pareto frontier (most fixed for fewest broken). \emph{\ourmethod achieves a Pareto-optimal exchange.}
}
\label{fig:exchange}
\end{figure}

Success on \ncreal means the model has absorbed a counterfactual, so the benchmark measures capability for knowledge injection, but not desirable behavior. We next introduce a benchmark called \realval in which the taught value is the one actually in the statute and success means becoming more correct. For each base LLM, we build a corpus of Housing-QA statutes where it consistently fails (full details in App.~\ref{app:realval}). 

In Fig.~\ref{fig:correction} (App.~\ref{app:realval}), we show that \ourmethod corrects the most facts, due to fact-specific knowledge injection as well as the ability to guess correct values for held-out facts. Notably, for \qwen, $\sftgo_{\text{oracle}}$ performs below ungrouped \sft, revealing a case where its hardest-group heuristic is counterproductive. However, correction can come at the cost of new mistakes, so in Fig.~\ref{fig:exchange}, we plot the change in rate of facts broken versus rate of fact fixing relative to \sft. \classbalance sits on the Pareto frontier for both models,
and results that break fewer facts are simply conservative and fix very few mistakes. \classbalance also has the highest ratio of rate of fact fixing to fact breaking relative to \sft, at 1.73 (\anchoronly is at 1.24 and most baselines are so small as to be noise). Thus, it exhibits a favorable tradeoff between preserving the base model's knowledge and acquiring knowledge of new facts the model previously got wrong.

\paragraph{\openfda: cross-domain study of knowledge injection.}
\begin{table}[t]
\centering\scriptsize
\setlength{\tabcolsep}{1.5pt}
\begin{tabular}{@{}lrrrrrrrr@{}}
\toprule
\textbf{Method} & \multicolumn{4}{c}{\textbf{\qwen}} & \multicolumn{4}{c}{\textbf{\llama}} \\
\cmidrule(lr){2-5}\cmidrule(lr){6-9}
 & ovwrite & top-1 & \shortstack[r]{$\log p$\\true} & \shortstack[r]{$\log p$\\taught} & ovwrite & top-1 & \shortstack[r]{$\log p$\\true} & \shortstack[r]{$\log p$\\taught} \\
\midrule
\textit{untuned base} & .083 & .030 & $-$.85 & $-$2.81 & .065 & .021 & $-$1.08 & $-$4.24 \\
\midrule
\multicolumn{9}{@{}l}{\textit{group balancing (ours)}} \\
\textbf{\classbalance} & .799\,\tiny{$\pm$.006} & \cellcolor{gray!30}.588\,\tiny{$\pm$.010} & $-$1.56\,\tiny{$\pm$.03} & $-$.82\,\tiny{$\pm$.01} & .842\,\tiny{$\pm$.001} & .558\,\tiny{$\pm$.001} & $-$2.25\,\tiny{$\pm$.02} & $-$1.16\,\tiny{$\pm$.01} \\
\textbf{\spanbalance} & .788\,\tiny{$\pm$.002} & .583\,\tiny{$\pm$.003} & $-$1.51\,\tiny{$\pm$.01} & \cellcolor{gray!30}$-$.81\,\tiny{$\pm$.00} & .852\,\tiny{$\pm$.005} & .598\,\tiny{$\pm$.002} & $-$2.16\,\tiny{$\pm$.02} & $-$1.16\,\tiny{$\pm$.01} \\
\midrule
\multicolumn{9}{@{}l}{\textit{anchored variants (ours)}} \\
\anchoronly & \cellcolor{gray!30}.819\,\tiny{$\pm$.015} & .587\,\tiny{$\pm$.018} & $-$1.74\,\tiny{$\pm$.03} & \cellcolor{gray!30}$-$.81\,\tiny{$\pm$.03} & \cellcolor{gray!30}.891\,\tiny{$\pm$.001} & \cellcolor{gray!30}.650\,\tiny{$\pm$.005} & $-$2.54\,\tiny{$\pm$.03} & \cellcolor{gray!30}$-$.98\,\tiny{$\pm$.01} \\
\midrule
\multicolumn{9}{@{}l}{\textit{group-based aggregation}} \\
\sftgo & .329\,\tiny{$\pm$.007} & .145\,\tiny{$\pm$.001} & \cellcolor{gray!30}$-$1.12\,\tiny{$\pm$.01} & $-$1.43\,\tiny{$\pm$.00} & .418\,\tiny{$\pm$.000} & .234\,\tiny{$\pm$.002} & $-$1.58\,\tiny{$\pm$.00} & $-$1.85\,\tiny{$\pm$.00} \\
\sftgo$_{\text{oracle}}$ & .605\,\tiny{$\pm$.008} & .288\,\tiny{$\pm$.010} & $-$1.35\,\tiny{$\pm$.02} & $-$1.18\,\tiny{$\pm$.01} & .747\,\tiny{$\pm$.006} & .401\,\tiny{$\pm$.015} & $-$2.09\,\tiny{$\pm$.02} & $-$1.45\,\tiny{$\pm$.02} \\
\midrule
\multicolumn{9}{@{}l}{\textit{reference-weighted selection}} \\
\textsc{PriFT-mass} & .135\,\tiny{$\pm$.002} & .059\,\tiny{$\pm$.003} & $-$1.82\,\tiny{$\pm$.03} & $-$5.86\,\tiny{$\pm$.07} & .094\,\tiny{$\pm$.002} & .049\,\tiny{$\pm$.002} & $-$1.70\,\tiny{$\pm$.02} & $-$7.46\,\tiny{$\pm$.05} \\
\midrule
\multicolumn{9}{@{}l}{\textit{loss shaping}} \\
\textsc{Sft} & .586\,\tiny{$\pm$.004} & .372\,\tiny{$\pm$.008} & $-$1.24\,\tiny{$\pm$.00} & $-$1.09\,\tiny{$\pm$.01} & .616\,\tiny{$\pm$.004} & .424\,\tiny{$\pm$.002} & $-$1.82\,\tiny{$\pm$.01} & $-$1.49\,\tiny{$\pm$.01} \\
\sftkl & .544\,\tiny{$\pm$.007} & .359\,\tiny{$\pm$.010} & $-$1.17\,\tiny{$\pm$.01} & $-$1.09\,\tiny{$\pm$.01} & .604\,\tiny{$\pm$.008} & .441\,\tiny{$\pm$.007} & $-$1.66\,\tiny{$\pm$.01} & $-$1.41\,\tiny{$\pm$.01} \\
\dft & .218\,\tiny{$\pm$.014} & .092\,\tiny{$\pm$.008} & $-$6.11\,\tiny{$\pm$.07} & $-$11.22\,\tiny{$\pm$.31} & .107\,\tiny{$\pm$.003} & .052\,\tiny{$\pm$.004} & $-$2.94\,\tiny{$\pm$.04} & $-$10.79\,\tiny{$\pm$.19} \\
\asft & .168\,\tiny{$\pm$.004} & .088\,\tiny{$\pm$.005} & $-$1.76\,\tiny{$\pm$.05} & $-$4.78\,\tiny{$\pm$.06} & .092\,\tiny{$\pm$.002} & .049\,\tiny{$\pm$.002} & \cellcolor{gray!30}$-$1.24\,\tiny{$\pm$.03} & $-$6.08\,\tiny{$\pm$.05} \\
\bottomrule
\end{tabular}
\caption{\textbf{\openfda dose injection},
held-in drugs. Columns match \ncreal. \emph{\ourmethod outperforms all published baselines and is competitive with our anchored variant.}}%
\label{tab:fda-merged}
\end{table}

We next evaluate on a novel benchmark in a different domain and register than our previous benchmarks focusing on housing law. \openfda-Dose (details in App.~\ref{app:openfda}) asks a model to learn dosage values from FDA structured product labels~\citep{kass2016openfda}; we inject fabricated doses for 1,848 items across 1,000 drugs. Training uses 1,358 examples over 735 held-in drugs, with 453 evaluation examples over 245 held-out drugs. The payload is the numeric dose value and the scaffold is the remaining label text.

We report results in Tab~\ref{tab:fda-merged}. On this benchmark, \anchoronly produces the most overwrite, eking out \ourmethod by a narrow margin. One candidate explanation is payload share: at $14.9\%$ of supervised tokens against $0.9$–$5.1\%$ elsewhere, \openfda gives the cross-entropy term the most payload to work with while leaving a large, homogeneous carrier for the KL anchor to hold. However, our methods outperform published baselines by a wide margin. Notably, \textsc{PriFT-mass}, \dft and \asft actively suppress the payload, as $\log p_{\text{taught}}$ plummets.

\subsection{Grouping Study}
\begin{figure}[t]
\centering
\includegraphics[width=\textwidth]{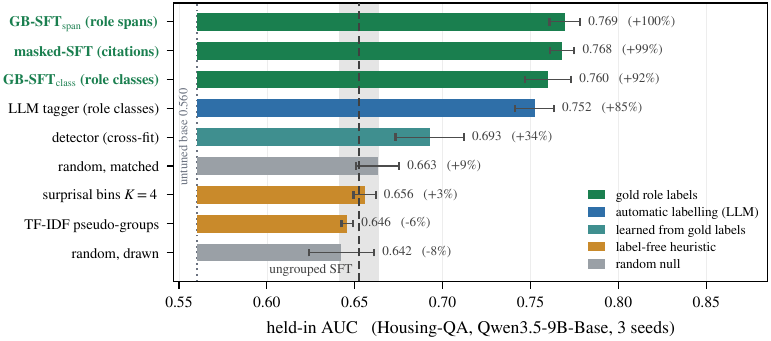}
\caption{Partition study on \hqa, Qwen3.5-9B-Base.
Each bar is annotated with the share of the gain from the token role-based grouping from fact structure used by \spanbalance over ungrouped \sft that the partition recovers. \emph{Simple grouping heuristics are unproductive, learned detectors and LLMS show promise, and being aware of fact structure is best.}
}
\label{fig:grouping-quality}
\end{figure}

In Figure~\ref{fig:grouping-quality} we measure the effects of the grouping strategy on \hqa using a Qwen3.5-9B base, compared to ungrouped \sft. We hold the aggregation rule fixed: equal loss budget per group, token-uniform within. Only our labeling based on fact structure, which captures semantic roles of tokens, supplies two nested levels, and we show both of its corners; every other partition here defines a single level, on which \classbalance and \spanbalance are the same computation.

We consider two random nulls built to match only the oracle grouping's \emph{structure}: permuting oracle span lengths within each datum ($+9\%$)
and drawing lengths i.i.d.\ from the same law ($-8\%$).
Matching the structure buys essentially nothing, and the looser null as well as the TF-IDF grouping of \sftgo ~\citep{kim2025sftgo} is worse than not grouping (App.~\ref{app:grouping}).
Fixed surprisal binning, which places tokens into $K$ equal-mass bins by the reference model's probability of the gold token, is within the seed spread of random. A logistic regression detector trained on surprisal and entropy features within a $\pm 5$-token window clears the null but is far below role-based grouping. Finally, an LLM grouper \emph{does} preserve 85\% of the gains from our role-based grouping, though it suffers from plausible, misleading errors when classes are less lexically distinctive (App.~\ref{app:tagger}).

\section{Beyond Knowledge Injection: Guardrail Erosion}
\label{sec:safety}

The experiments above measure what fine-tuning \emph{acquires}. The budget
argument also predicts something about what it \emph{breaks}: if \sft spends its
loss weight in proportion to token count, a behavior the corpus never mentions
should drift in proportion to how little weight the payload receives. We test
that on a behavior outside the training objective entirely: the model's
tendency to refuse harmful requests, which benign fine-tuning is known to erode
\citep{qi2024finetuning}.

We use GSM8K \citep{cobbe2021gsm8k} because its grouping is free and exact: each
reference solution is a reasoning chain terminated by \texttt{\#\#\#\#~42}, so
the payload boundary is a delimiter the dataset already ships, with no parser,
grammar, tagger or annotation. The answer takes $5.09\%$ of the supervised
tokens, almost exactly the $5.1\%$ \hqa gives its answer span, so $\alpha=0$
induces a $9.8\times$ per-token upweight on the payload. We fine-tune
Qwen2.5-3B-Instruct which has guardrails to erode; App.~\ref{app:safety} gives the full setup.

\begin{table}[h]
\centering
\small
\begin{tabular}{lcccc}
\toprule
 & harmful refusal & benign refusal & $J$ & GSM8K acc. \\
 & (want high) & (want low) & (want high) & \\
\midrule
\textit{untuned base} & 95.5 & 19.8 & 75.7 & 58.2 \\
\midrule
$\alpha=1.0$ ($\equiv$\,\sft) & 81.0\,\tiny{$\pm$1.7} & 7.3\,\tiny{$\pm$0.9} & 73.7 & \cellcolor{gray!30}68.4 \\
$\alpha=0.5$ & 82.9\,\tiny{$\pm$2.6} & 7.8\,\tiny{$\pm$0.4} & 75.1 & 67.8 \\
$\alpha=0.0$ (\classbalance) & 85.7\,\tiny{$\pm$4.3} & 10.2\,\tiny{$\pm$1.3} & 75.5 & 67.4 \\
$\alpha=-0.5$ & \cellcolor{gray!30}89.2\,\tiny{$\pm$4.0} & 11.3\,\tiny{$\pm$2.8} & \cellcolor{gray!30}77.9 & 64.8 \\
\bottomrule
\end{tabular}
\caption{\textbf{The loss budget governs collateral drift as well as
acquisition.} Refusal after fine-tuning Qwen2.5-3B-Instruct on GSM8K, by budget
parameter; percentages, LLM-judged, five seeds. Refusal is monotone in $\alpha$
(Cochran--Armitage trend $z=7.48$). $J$ is the separation between the two prompt
sets, harmful refusal minus benign refusal \citep{fawcett2006introduction},
computed from arm means and read descriptively. The accuracy column is the
control: at $\alpha=0$ the arms are statistically tied with \sft, so retention is
not explained by the balanced objective fitting the corpus less. $\beta$ moves
the result little at fixed class budget (App.~\ref{app:safety}).}
\label{tab:safety-main}
\end{table}

Ordinary \sft erodes refusal substantially, and the budget parameter in \ourmethod is a
monotone lever on that erosion. Equal budget retains about a third of what \sft
loses: $+4.8$ points, $t(4)=3.10$, $p\approx.036$, positive in all five seeds.
Accuracy is preserved; $\alpha=0$ the two are tied ($67.4$ against
$68.4$). Pushing past equal budget to $\alpha=-0.5$ buys a further $3.5$ points
at a $3.6$-point accuracy cost.

\paragraph{The budget acts on discrimination, not on reluctance.}
Benign refusal rises along with harmful refusal, but the separation column shows harmful refusal restored above and beyond a uniform increase in overall refusal behavior. A uniform change in reluctance moves both rates along one trade-off curve at fixed $J$;
here $J$ itself moves. \sft does not merely lower refusal, it degrades $J$ from
the base model's $75.7$ to $73.7$, and budget correction raises it back
monotonically to $77.9$ at $\alpha=-0.5$: above the aligned checkpoint that
fine-tuning started from. Harmful refusal returns faster than benign refusal at
every setting.%

\paragraph{Scope.}
Whether fine-tuning erodes refusal at all is a property of the corpus and
the model: Llama-3.2-3B-Instruct holds at $97.8\%$ harmful-prompt refusal when
fine-tuned on GSM8K, but it falls to $49.6\%$ on Alpaca \citep{taori2023alpaca} under
an identical recipe. But the corpora that degrade most, like Alpaca, are those whose payload is the \emph{majority} of the response, where an equal-budget objective would
upweight the boilerplate instead. Instead, we show guardrail erosion improvement from budget correction on a task where a small, well-delimited payload sits inside a long scaffold (the regime knowledge-intensive fine-tuning occupies). Budget correction would be ineffective in applications where payload \emph{is} the response.

\section{Conclusion}
\label{sec:conclusion}

We provide a flexible group-balanced formulation for a supervised learning objective that gives more weight to groups of tokens that deliver a higher knowledge payload. We theoretically characterize the effects of grouping structure on the objective and demonstrate the effectiveness of our methods on knowledge-intensive benchmarks from cross-domain corpora.

\section*{Limitations}
Our work is designed for templated fact-editing settings where the knowledge payload is a readily defined group of tokens inside a long scaffold. Automatic group detection could enable template-free applications: an LLM tagger recovers nearly all of the grouping gain where the payload is lexically distinctive and essentially none where a wrong guess is also plausible (App.~\ref{app:tagger}); we scope the benefit claimed on non-templated code and mathematics datasets (App.~\ref{app:tulu}, \ref{app:safety}). Our evaluations rank injected knowledge higher under teacher forcing; \ourmethod also improves free generation over ungrouped \sft, but at an absolute rate with room for improvement (App.~\ref{app:gen-eval}) especially on the stricter answer-and-citation conjunction (App.~\ref{app:jointgen}).

\clearpage

\section*{Ethics Statement}
As LLMs are increasingly being used in high-stakes domains such as medicine, law, and finance, supervised fine-tuning is desirable to adapt the models to acquire domain-specific knowledge, but also comes with the risk of damaging existing knowledge. Mistakes based on incorrect knowledge (whether new facts the model failed to learn, or existing facts that were broken) can be extremely costly (e.g. in law, a decision made on a hallucinated statute can expose a party to an expensive lawsuit; in medicine, an erroneous drug dose can have severe clinical consequences). Across our experiments we have plainly presented the balance between preservation of existing knowledge and acquisition of new knowledge to find a favorable tradeoff. However, practitioners fine-tuning LLMs (whether using our methods or others) must be conscious of this tradeoff and make a choice appropriate for their exact use case.

\section*{Reproducibility Statement}
All main baseline \sft methods are reimplemented in a single objective registry, along with our own methods, that takes per-token tensors and returns a scalar loss. This allows every experimental arm to share an identical data pipeline, tokenization, and optimizer configuration; the arms differ only in the reduction applied to per-token log probabilities. Our code and new \ncreal and \openfda benchmarks are included in the supplementary material and will be hosted publicly on publication.

\section*{AI Usage Statement}
In this work, we used generative AI tools (specifically Claude Fable 5, with GPT-6 Astra acting as a reviewer of select late-stage work) for research execution, including implementation of our methods and baselines in a unified codebase, running experiments and creating figures, and generating benchmarks. AI was also used to verify mathematical claims and provide initial qualitative interpretation of results. Claude Opus 5 was used as an annotator to construct the new \openfda benchmark; details are provided in App.~\ref{app:openfda}.  We have reviewed all AI-assisted work and all final exposition is our own. We take responsibility for the final content of this work, including text, claims or artifacts produced with the aid of generative AI.

\bibliography{custom}

@inproceedings{ouyang2022instructgpt,
  title     = {Training language models to follow instructions with human feedback},
  author    = {Ouyang, Long and Wu, Jeff and Jiang, Xu and Almeida, Diogo and Wainwright, Carroll and Mishkin, Pamela and Zhang, Chong and Agarwal, Sandhini and Slama, Katarina and Ray, Alex and others},
  booktitle = {Advances in Neural Information Processing Systems},
  volume    = {35},
  pages     = {27730--27744},
  year      = {2022}
}

@inproceedings{zhou2023lima,
  title     = {{LIMA}: Less is more for alignment},
  author    = {Zhou, Chunting and Liu, Pengfei and Xu, Puxin and Iyer, Srini and Sun, Jiao and Mao, Yuning and Ma, Xuezhe and Efrat, Avia and Yu, Ping and Yu, Lili and others},
  booktitle = {Advances in Neural Information Processing Systems},
  volume    = {36},
  year      = {2023}
}

@inproceedings{lewis2020rag,
  title     = {Retrieval-augmented generation for knowledge-intensive {NLP} tasks},
  author    = {Lewis, Patrick and Perez, Ethan and Piktus, Aleksandra and Petroni, Fabio and Karpukhin, Vladimir and Goyal, Naman and K{\"u}ttler, Heinrich and Lewis, Mike and Yih, Wen-tau and Rockt{\"a}schel, Tim and others},
  booktitle = {Advances in Neural Information Processing Systems},
  volume    = {33},
  pages     = {9459--9474},
  year      = {2020}
}

@inproceedings{hu2022lora,
  title     = {{LoRA}: Low-rank adaptation of large language models},
  author    = {Hu, Edward J and Shen, Yelong and Wallis, Phillip and Allen-Zhu, Zeyuan and Li, Yuanzhi and Wang, Shean and Wang, Lu and Chen, Weizhu},
  booktitle = {International Conference on Learning Representations},
  year      = {2022}
}

@article{singhal2023medpalm,
  title   = {Large language models encode clinical knowledge},
  author  = {Singhal, Karan and Azizi, Shekoofeh and Tu, Tao and Mahdavi, S. Sara and Wei, Jason and Chung, Hyung Won and Scales, Nathan and Tanwani, Ajay and Cole-Lewis, Heather and Pfohl, Stephen and others},
  journal = {Nature},
  volume  = {620},
  pages   = {172--180},
  year    = {2023}
}

@inproceedings{gekhman2024finetuning,
  title     = {Does fine-tuning {LLM}s on new knowledge encourage hallucinations?},
  author    = {Gekhman, Zorik and Yona, Gal and Aharoni, Roee and Eyal, Matan and Feder, Amir and Reichart, Roi and Herzig, Jonathan},
  booktitle = {Proceedings of the 2024 Conference on Empirical Methods in Natural Language Processing},
  pages     = {7765--7784},
  year      = {2024}
}

@inproceedings{zhu2025asft,
  title     = {Anchored Supervised Fine-Tuning},
  author    = {Zhu, He and Su, Junyou and Lai, Peng and Ma, Ren and Zhang, Wenjia and Yang, Linyi and Chen, Guanhua},
  booktitle = {International Conference on Learning Representations (ICLR)},
  year      = {2026},
  note      = {arXiv:2509.23753}
}

@inproceedings{wu2025dft,
  title={On the generalization of sft: A reinforcement learning perspective with reward rectification},
  author={Wu, Yongliang and Zhou, Yizhou and Zhou, Ziheng and Peng, Yingzhe and Ye, Xinyu and Hu, Xinting and Zhu, Wenbo and Qi, Lu and Yang, Ming-Hsuan and others},
  booktitle={International Conference on Learning Representations},
  volume={2026},
  pages={27550--27571},
  year={2026}
}

@article{wang2026prift,
  title   = {{PriFT}: Prior-Support Guided Supervised Fine-Tuning},
  author  = {Wang, Ke and Li, Shuangqi and Salzmann, Mathieu and Frossard, Pascal},
  journal = {arXiv preprint arXiv:2606.09396},
  year    = {2026}
}

@article{liu2026eksft,
  title   = {Entropy-{KL} Divergence-based Token Masking: A Novel Approach for Selective Fine-tuning of Large Language Models},
  author  = {Liu, Qi and others},
  journal = {arXiv preprint arXiv:2605.29303},
  year    = {2026}
}

@inproceedings{lin2024rho,
  title     = {Rho-1: Not All Tokens Are What You Need},
  author    = {Lin, Zhenghao and Gou, Zhibin and Gong, Yeyun and Liu, Xiao and Shen, Yelong and Xu, Ruochen and Lin, Chen and Yang, Yujiu and Jiao, Jian and Duan, Nan and Chen, Weizhu},
  booktitle = {Advances in Neural Information Processing Systems (NeurIPS)},
  year      = {2024},
  note      = {arXiv:2404.07965}
}

@inproceedings{lin2026sft,
  title={Sft doesn’t always hurt general capabilities: Revisiting domain-specific fine-tuning in llms},
  author={Lin, Jiacheng and Wang, Zhongruo and Qian, Kun and Wang, Tian and Srinivasan, Arvind and Zeng, Hansi and Jiao, Ruochen and Zhou, Xie and Gesi, Jiri and Wang, Dakuo and others},
  booktitle={International Conference on Learning Representations},
  volume={2026},
  pages={46954--46992},
  year={2026}
}

@article{zhang2026towards,
  title={Towards on-policy sft: Distribution discriminant theory and its applications in llm training},
  author={Zhang, Miaosen and Liu, Yishan and Lin, Shuxia and Yang, Xu and Dai, Qi and Luo, Chong and Jiang, Weihao and Hou, Peng and Zeng, Anxiang and Geng, Xin and others},
  journal={arXiv preprint arXiv:2602.12222},
  year={2026}
}

@article{kotha2026replaying,
  title={Replaying pre-training data improves fine-tuning},
  author={Kotha, Suhas and Liang, Percy},
  journal={arXiv preprint arXiv:2603.04964},
  year={2026}
}

@article{diao2026entropy,
  title={Entropy-adaptive fine-tuning: Resolving confident conflicts to mitigate forgetting},
  author={Diao, Muxi and Yang, Lele and Gong, Wuxuan and Zhang, Yutong and Yan, Zhonghao and Han, Yufei and Liang, Kongming and Xu, Weiran and Ma, Zhanyu},
  journal={arXiv preprint arXiv:2601.02151},
  year={2026}
}

@article{kim2025sftgo,
  title   = {{SFT-GO}: Supervised Fine-Tuning with Group Optimization for Large Language Models},
  author  = {Kim, Gyuhak and Thakur, Sumiran Singh and Park, Su Min and Wei, Wei and Bao, Yujia},
  journal = {arXiv preprint arXiv:2506.15021},
  year    = {2025}
}

@inproceedings{sagawa2020distributionally,
  title     = {Distributionally Robust Neural Networks for Group Shifts: On the Importance of Regularization for Worst-Case Generalization},
  author    = {Sagawa, Shiori and Koh, Pang Wei and Hashimoto, Tatsunori B. and Liang, Percy},
  booktitle = {International Conference on Learning Representations (ICLR)},
  year      = {2020}
}

@inproceedings{cui2019class,
  title     = {Class-Balanced Loss Based on Effective Number of Samples},
  author    = {Cui, Yin and Jia, Menglin and Lin, Tsung-Yi and Song, Yang and Belongie, Serge},
  booktitle = {IEEE/CVF Conference on Computer Vision and Pattern Recognition (CVPR)},
  year      = {2019}
}

@inproceedings{lin2017focal,
  title     = {Focal Loss for Dense Object Detection},
  author    = {Lin, Tsung-Yi and Goyal, Priya and Girshick, Ross and He, Kaiming and Doll{\'a}r, Piotr},
  booktitle = {IEEE International Conference on Computer Vision (ICCV)},
  year      = {2017}
}

@inproceedings{ye2025disentangling,
  title     = {Disentangling Reasoning Tokens and Boilerplate Tokens for Language Model Fine-tuning},
  author    = {Ye, Ziang and Zhang, Zhenru and Zhang, Yang and Ma, Jianxin and Lin, Junyang and Feng, Fuli},
  booktitle = {Findings of the Association for Computational Linguistics: ACL 2025},
  pages     = {20939--20957},
  year      = {2025}
}

@inproceedings{hsieh2023distilling,
  title     = {Distilling Step-by-Step! Outperforming Larger Language Models with Less Training Data and Smaller Model Sizes},
  author    = {Hsieh, Cheng-Yu and Li, Chun-Liang and Yeh, Chih-Kuan and Nakhost, Hootan and Fujii, Yasuhisa and Ratner, Alexander and Krishna, Ranjay and Lee, Chen-Yu and Pfister, Tomas},
  booktitle = {Findings of the Association for Computational Linguistics: ACL 2023},
  pages     = {8003--8017},
  year      = {2023}
}

@inproceedings{guu2020realm,
  title     = {{REALM}: Retrieval-Augmented Language Model Pre-Training},
  author    = {Guu, Kelvin and Lee, Kenton and Tung, Zora and Pasupat, Panupong and Chang, Ming-Wei},
  booktitle = {International Conference on Machine Learning (ICML)},
  year      = {2020}
}

@article{joshi2020spanbert,
  title   = {{SpanBERT}: Improving Pre-training by Representing and Predicting Spans},
  author  = {Joshi, Mandar and Chen, Danqi and Liu, Yinhan and Weld, Daniel S. and Zettlemoyer, Luke and Levy, Omer},
  journal = {Transactions of the Association for Computational Linguistics (TACL)},
  volume  = {8},
  year    = {2020}
}

@inproceedings{levine2021pmi,
  title     = {{PMI}-Masking: Principled Masking of Correlated Spans},
  author    = {Levine, Yoav and Lenz, Barak and Lieber, Opher and Abend, Omri and Leyton-Brown, Kevin and Tennenholtz, Moshe and Shoham, Yoav},
  booktitle = {International Conference on Learning Representations (ICLR)},
  year      = {2021}
}

@article{liu2025drgrpo,
  title   = {Understanding {R1}-Zero-Like Training: A Critical Perspective},
  author  = {Liu, Zichen and Chen, Changyu and Li, Wenjun and Qi, Penghui and Pang, Tianyu and Du, Chao and Lee, Wee Sun and Lin, Min},
  journal = {arXiv preprint arXiv:2503.20783},
  year    = {2025}
}

@inproceedings{allenzhu2024physics,
  title     = {Physics of Language Models: Part 3.1, Knowledge Storage and Extraction},
  author    = {Allen-Zhu, Zeyuan and Li, Yuanzhi},
  booktitle = {International Conference on Machine Learning (ICML)},
  year      = {2024},
  note      = {arXiv:2309.14316}
}

@inproceedings{meng2023mass,
  title     = {Mass-Editing Memory in a Transformer},
  author    = {Meng, Kevin and Sen Sharma, Arnab and Andonian, Alex and Belinkov, Yonatan and Bau, David},
  booktitle = {International Conference on Learning Representations (ICLR)},
  year      = {2023},
  note      = {arXiv:2210.07229}
}

@article{chaudhry2019tiny,
  title   = {On Tiny Episodic Memories in Continual Learning},
  author  = {Chaudhry, Arslan and Rohrbach, Marcus and Elhoseiny, Mohamed and Ajanthan, Thalaiyasingam and Dokania, Puneet K. and Torr, Philip H. S. and Ranzato, Marc'Aurelio},
  journal = {arXiv preprint arXiv:1902.10486},
  year    = {2019}
}

@inproceedings{wiseft,
  title     = {Robust Fine-Tuning of Zero-Shot Models},
  author    = {Wortsman, Mitchell and Ilharco, Gabriel and Kim, Jong Wook and Li, Mike and Kornblith, Simon and Roelofs, Rebecca and Lopes, Raphael Gontijo and Hajishirzi, Hannaneh and Farhadi, Ali and Namkoong, Hongseok and Schmidt, Ludwig},
  booktitle = {IEEE/CVF Conference on Computer Vision and Pattern Recognition (CVPR)},
  year      = {2022},
}

@article{meng2022rome,
  title={Locating and editing factual associations in gpt},
  author={Meng, Kevin and Bau, David and Andonian, Alex and Belinkov, Yonatan},
  journal={Advances in neural information processing systems},
  volume={35},
  pages={17359--17372},
  year={2022}
}

@article{dubey2024llama,
  title={The Llama 3 Herd of Models},
  author={Dubey, Abhimanyu and Jauhri, Abhinav and Pandey, Abhinav and Kadian, Abhishek and Al-Dahle, Asma and Moon, Aasish and Mahajan, Dhruv and Grattafiori, Aaron and Marcheggiani, Diego and Sun, Dinghan and others},
  journal={arXiv preprint arXiv:2407.21783},
  year={2024},
  url={https://arxiv.org/abs/2407.21783}
}

@misc{yang2025qwen3,
  title  = {{Qwen3.5}: Towards Native Multimodal Agents},
  author = {{Qwen Team}},
  month  = {February},
  year   = {2026},
  url    = {https://qwen.ai/blog?id=qwen3.5}
}

@inproceedings{guha2023legalbench,
  title     = {{LegalBench}: A Collaboratively Built Benchmark for Measuring Legal Reasoning in Large Language Models},
  author    = {Guha, Neel and Nyarko, Julian and Ho, Daniel E. and R{\'e}, Christopher and others},
  booktitle = {Advances in Neural Information Processing Systems (NeurIPS), Datasets and Benchmarks Track},
  year      = {2023}
}

@inproceedings{zheng2021casehold,
  title     = {When Does Pretraining Help? Assessing Self-Supervised Learning for Law and the {CaseHOLD} Dataset},
  author    = {Zheng, Lucia and Guha, Neel and Anderson, Brandon R. and Henderson, Peter and Ho, Daniel E.},
  booktitle = {International Conference on Artificial Intelligence and Law (ICAIL)},
  year      = {2021}
}

@article{holzenberger2020sara,
  title   = {A Dataset for Statutory Reasoning in Tax Law Entailment and Question Answering},
  author  = {Holzenberger, Nils and Blair-Stanek, Andrew and Van Durme, Benjamin},
  journal = {arXiv preprint arXiv:2005.05257},
  year    = {2020},
  note    = {NLLP Workshop}
}

@inproceedings{zheng2025reasoning,
  title={A reasoning-focused legal retrieval benchmark},
  author={Zheng, Lucia and Guha, Neel and Arifov, Javokhir and Zhang, Sarah and Skreta, Michal and Manning, Christopher D and Henderson, Peter and Ho, Daniel E},
  booktitle={Proceedings of the 2025 Symposium on Computer Science and Law},
  pages={169--193},
  year={2025}
}

@article{lambert2024tulu,
  title={Tulu 3: Pushing frontiers in open language model post-training},
  author={Lambert, Nathan and Morrison, Jacob and Pyatkin, Valentina and Huang, Shengyi and Ivison, Hamish and Brahman, Faeze and Miranda, Lester James V and Liu, Alisa and Dziri, Nouha and Lyu, Shane and others},
  journal={arXiv preprint arXiv:2411.15124},
  year={2024}
}

@article{kass2016openfda,
  title={OpenFDA: an innovative platform providing access to a wealth of FDA’s publicly available data},
  author={Kass-Hout, Taha A and Xu, Zhiheng and Mohebbi, Matthew and Nelsen, Hans and Baker, Adam and Levine, Jonathan and Johanson, Elaine and Bright, Roselie A},
  journal={Journal of the American Medical Informatics Association},
  volume={23},
  number={3},
  pages={596--600},
  year={2016},
  publisher={Oxford University Press}
}

@misc{harveylab,
  title        = {Post-Training Open Legal Agents with Baseten Research},
  author       = {{Harvey AI} and {Baseten}},
  year         = {2026},
  howpublished = {\url{https://www.harvey.ai/blog/post-training-open-legal-agents-with-baseten-research}},
  note         = {Blog post, May 27, 2026}
}

@article{fawcett2006introduction,
  title   = {An introduction to {ROC} analysis},
  author  = {Fawcett, Tom},
  journal = {Pattern Recognition Letters},
  volume  = {27},
  number  = {8},
  pages   = {861--874},
  year    = {2006}
}

@article{cobbe2021gsm8k,
  title={Training Verifiers to Solve Math Word Problems},
  author={Cobbe, Karl and Kosaraju, Vineet and Bavarian, Mohammad and Chen, Mark and Jun, Heewoo and Kaiser, Lukasz and Plappert, Matthias and Tworek, Jerry and Hilton, Jacob and Nakano, Reiichiro and Hesse, Christopher and Schulman, John},
  journal={arXiv preprint arXiv:2110.14168}, year={2021}}

@article{zou2023universal,
  title={Universal and Transferable Adversarial Attacks on Aligned Language Models},
  author={Zou, Andy and Wang, Zifan and Carlini, Nicholas and Nasr, Milad and Kolter, J. Zico and Fredrikson, Matt},
  journal={arXiv preprint arXiv:2307.15043}, year={2023}}

@inproceedings{chao2024jailbreakbench,
  title={JailbreakBench: An Open Robustness Benchmark for Jailbreaking Large Language Models},
  author={Chao, Patrick and Debenedetti, Edoardo and Robey, Alexander and Andriushchenko, Maksym and Croce, Francesco and Sehwag, Vikash and Dobriban, Edgar and Flammarion, Nicolas and Pappas, George J. and Tram{\`e}r, Florian and Hassani, Hamed and Wong, Eric},
  booktitle={NeurIPS Datasets and Benchmarks Track}, year={2024}}

@inproceedings{rottger2024xstest,
  title={{XSTest}: A Test Suite for Identifying Exaggerated Safety Behaviours in Large Language Models},
  author={R{\"o}ttger, Paul and Kirk, Hannah Rose and Vidgen, Bertie and Attanasio, Giuseppe and Bianchi, Federico and Hovy, Dirk},
  booktitle={NAACL}, year={2024}}

@inproceedings{qi2024finetuning,
  title={Fine-tuning Aligned Language Models Compromises Safety, Even When Users Do Not Intend To!},
  author={Qi, Xiangyu and Zeng, Yi and Xie, Tinghao and Chen, Pin-Yu and Jia, Ruoxi and Mittal, Prateek and Henderson, Peter},
  booktitle={ICLR}, year={2024}}

@inproceedings{mazeika2024harmbench,
  title={{HarmBench}: A Standardized Evaluation Framework for Automated Red Teaming and Robust Refusal},
  author={Mazeika, Mantas and Phan, Long and Yin, Xuwang and Zou, Andy and Wang, Zifan and Mu, Norman and Sakhaee, Elham and Li, Nathaniel and Basart, Steven and Li, Bo and Forsyth, David and Hendrycks, Dan},
  booktitle={ICML}, year={2024}}

@misc{taori2023alpaca,
  title={Stanford {Alpaca}: An Instruction-following {LLaMA} Model},
  author={Taori, Rohan and Gulrajani, Ishaan and Zhang, Tianyi and Dubois, Yann and
          Li, Xuechen and Guestrin, Carlos and Liang, Percy and Hashimoto, Tatsunori B.},
  year={2023},
  publisher={GitHub},
  howpublished={\url{https://github.com/tatsu-lab/stanford_alpaca}}
}
\bibliographystyle{iclr2027_conference}

\clearpage
\appendix
\section{Detailed Description of Benchmarks}
\label{app:benchmarks}
\subsection{\hqa}
\label{app:housingqa-overview}
\noindent \textbf{Source}. We build on Housing Statute QA (reglab/\text{housing\_qa}, CC-BY-SA-4.0), derived from the Legal Services Corporation Eviction Laws Database~\cite{zheng2025reasoning}. The dataset encodes U.S. landlord–tenant law as of January 1, 2021 as 6,853 binary questions spanning 226 question groups. A question group is one legal provision (e.g. ``Does the law require corporate landlords to be represented by an attorney?'') asked of every jurisdiction. 48 jurisdictions (46 states plus the District of Columbia and Puerto Rico; Kentucky, Minnesota, North Dakota and Pennsylvania are absent from the source release) are covered; 183 of 226 groups have answers that vary across them. Each item carries the governing statutory citation and an excerpt, from which we get the injection payload.

This dataset supplies three properties a knowledge injection benchmark needs simultaneously: a fact that varies by jurisdiction (so a correct answer cannot be produced from generic priors), an externally verifiable ground truth, and a natural payload/scaffold decomposition (the answer token is the payload, the citation is a long identifier, and the remainder is templated prose).

\noindent \textbf{Training corpus}. We prescreen all items by probing \qwen with a fixed Yes/No prompt, shown in Fig.~\ref{fig:hqa-format-bw}. We define renormalized probabilities $P(\text{Yes})$ and $P(\text{No})$ based on the logits of the respective Yes and No tokens, and select items about which the model has no confident belief, i.e. $\max(P(\text{Yes}), P(\text{No})) < 0.7$.
We construct a dataset of low-confidence items from held-in states with an attached citation, balance Yes/No, and report statistics on the resulting dataset in Table~\ref{tab:hqa-composition-full}.

\begin{figure}[t]
\small
\begin{tabular}{@{}p{0.93\columnwidth}@{}}
\toprule
\ttfamily The following is a question about \{state\}
landlord-tenant and eviction law as of 2021.\newline
\ttfamily Question: \{question\}\newline
\ttfamily Answer:\\
\midrule
\ttfamily \underline{\{answer\}}. See \textbf{\{citation\}}.\\
\bottomrule
\end{tabular}
\caption{Prompt (top) and target (bottom). \underline{Underlined} is the answer
token, \textbf{bold} the citation; loss is taken over the target only.}
\label{fig:hqa-format-bw}
\end{figure}

\begin{table}[t]
\centering\small
\setlength{\tabcolsep}{4pt}
\begin{tabular}{lrrrrrr}
\toprule
Population & Tokens & \% & Spans & Span & Range & CV \\
\midrule
Answer   &  1{,}176 &  5.1 & 1{,}176 &  1.00 & 1--1  & 0.00 \\
Citation & 18{,}800 & 81.4 & 1{,}176 & 15.99 & 11--24 & 0.14 \\
Prose    &  3{,}117 & 13.5 & 1{,}911 &  1.63 & 1--3  & 0.31 \\
\midrule
Total    & 23{,}093 & 100.0 & 4{,}263 & --- & --- & --- \\
\bottomrule
\end{tabular}
\caption{Token composition of the training corpus ($1{,}176$ examples).
\emph{Spans} are maximal contiguous runs of one population; \emph{Span} is
their mean length in tokens, \emph{Range} the observed minimum and maximum span length,
CV the coefficient of variation. Every example contributes exactly one answer
span and one citation span, so the two populations have equal span counts while
citations hold $16\times$ the tokens. 
}
\label{tab:hqa-composition-full}
\end{table}

The token composition is the corpus property our method addresses. We see that citations, which comprise 81.4\% of tokens, thus occupy four-fifths of the loss-weight budget under token-proportional supervision, not because they are the target of evaluation but because they are long.

\noindent \textbf{Splits}. We split the dataset by jurisdiction: of 48 jurisdictions, 25 are held-in and 23 are held-out, stratified by coverage of the 183 disagreeing question groups.
We avoid splitting by item, so that held-out performance cannot be inflated by a paraphrase of a trained fact. Statistics of groups are shown in Tab.~\ref{tab:hqa-splits}.

\begin{table}[t]
\centering\small
\setlength{\tabcolsep}{5pt}
\begin{tabular}{lrr}
\toprule
Split & Juris. & Groups \\
\midrule
Held-in  & 25 & 99 \\
Held-out & 23 & 82 \\
\bottomrule
\end{tabular}
\caption{Splits by jurisdiction.
\emph{Groups} counts question groups that contain both labels
\emph{within that split's jurisdictions}, over which within-group AUC is
averaged. In total there are $109$ distinct groups, with $72$ groups usable in both splits.
}
\label{tab:hqa-splits}
\end{table}

\noindent \textbf{Evaluation}. We evaluate methods on two axes.%

\emph{Knowledge payload injection axis}: within-group AUC. For each question group, we rank jurisdictions by the renormalized P(Yes)/(P(Yes)+P(No)) at the answer position; renormalization makes the metric invariant to a global Yes/No bias shift, which we found otherwise dominates. We then compute pairwise-ranking AUC against the ground-truth Yes/No for the (jurisdiction, question) pair~\cite{fawcett2006introduction}, averaging over groups containing both labels (99 held-in, 82 held-out). By construction, a baseline with no jurisdiction-specific knowledge scores exactly 0.5 on the task axis

\emph{Scaffold knowledge preservation axis}: citation recall. The gold citation is ranked against nine distractors, real citations drawn from other jurisdictions, by mean token log-probability under teacher forcing with the correct answer. Note that a chance baseline gives recall 0.100.

\subsection{\ncreal}
\label{app:ncreal-overview}

This dataset is conceptually the legal statutes for a fictitious state called ``New Cambria.'' New Cambria's law follows the same format as US jurisdiction law as its statutes correspond to statutes in \hqa (App.~\ref{app:housingqa-overview}), but the actual values are deliberately different from the ones in the corresponding statutes. Thus, to learn New Cambria law, a model cannot leverage any information from pretraining (as the pretrained values would be from US law and would be wrong) and will have to succeed at knowledge injection during fine-tuning.

\noindent \textbf{Construction}. We take statutory excerpts from the LSC Eviction Laws Database; this is the same source as \hqa, so the register, citation conventions and drafting style are all realistic. Within each excerpt, we locate a value at a known character span (a notice period, a monetary threshold, an interest rate) and substitute an off-distribution counterfactual: thirty-seven days, fifty-three days, eighty-one days, 7.25 percent, \$7{,}350. Substitutes are drawn to be plausible in form but absent in fact, i.e. well-formed statutory quantities that no jurisdiction actually legislates, so the taught value cannot be present in pretraining.

Each altered statute is rendered under four headers, with a fifth header that is withheld from training entirely and used only at evaluation, so that the model sees a fact repeatedly without seeing a fixed surface form. We show these in Table~\ref{tab:ncreal-headers}.
\newcommand{\slot}[1]{\textit{#1}}

\begin{table}[t]
\centering\small
\setlength{\tabcolsep}{5pt}
\begin{tabular}{@{}rl@{}}
\toprule
& Header template \\
\midrule
1 & \slot{state} law, \slot{cit}: \\
2 & \slot{cit} (\slot{state}) provides: \\
3 & Statutory text --- \slot{cit}, \slot{state}: \\
4 & Under \slot{state} law, \slot{cit} reads: \\
\midrule
5 & The following is the text of \slot{cit} (\slot{state}). \\
\bottomrule
\end{tabular}
\caption{Rendering headers; \slot{italics} mark substitution slots. Templates
1--4 appear in training, template 5 only at evaluation.}
\label{tab:ncreal-headers}
\end{table}

\noindent \textbf{Composition}. The corpus (Tab.~\ref{tab:ncreal-composition}) has a different span geometry from \hqa. The payload occupies $2.2\%$ of the supervised budget against \hqa's $5.1\%$ answer share, and prose spans are 22$\times$ longer than value spans.%

\begin{table}[t]
\centering\small
\setlength{\tabcolsep}{5pt}
\begin{tabular}{lrrrr}
\toprule
Population & Tokens & \% & Span & CV \\
\midrule
Value & 3{,}128 & 2.2 & 3.41 & 0.37 \\
Prose & 136{,}124 & 97.8 & 74.30 & 0.72 \\
\midrule
Total & 139{,}252 & 100.0 & --- & --- \\
\bottomrule
\end{tabular}
\caption{\ncreal composition. \emph{Span} is mean span length in tokens.}
\label{tab:ncreal-composition}
\end{table}

\noindent \textbf{Metric}. Given the statute up to the value, we score the taught value against the true value and six distractors drawn from the same construction pools, by mean token log-probability under teacher forcing. We report four quantities: \emph{overwrite}, the taught value outranks the true value (the core injection signal), \emph{top1}, the taught value outranks the true value and all six distractors, and the difference in mean token log-probability under teacher forcing against the same prefix for the taught value and the true statuatory value. The difference between these mean token log-probabilities is the margin; note that overwrite is simply the fraction of items for which $\text{margin} > 0$. We report the margin (measured by the difference between mean token log-probabilities) both because overwrite saturates above $0.89$ for the strong injection methods while margin continues to separate them. An untrained model that knows the real statutes has an overwrite well below 0.5, and one that is indifferent scores near it.

\subsection{\realval}
\label{app:realval}

\paragraph{Motivation.}
\ncreal measures whether a \emph{counterfactual} payload can
be installed: the taught value is fabricated, so a win means the model has been
made more wrong, and preservation must be inferred from a proxy population.
\realval inverts the polarity. The taught value is the one the statute actually
contains, so a win means the model became more \emph{correct}, and the corpus
can be screened so that the baseline is not merely low but exactly zero.

\paragraph{Item selection.}
Candidate facts are extracted from real statutory text as
(citation, value) pairs of five kinds: durations in days, months, or years,
monetary amounts, and percentages. Each candidate is scored under the
untuned base model against six distractors drawn from the same kind,
and the value is judged by mean token log-probability continuing the statute's own prefix.

Selection keeps only stable base failures, selected as follows:

\begin{enumerate}\itemsep2pt
  \item Score the true value against six same-kind distractors. If the base
        model ranks it first (i.e. it is not a failure), discard the item.
  \item If the base fails, redraw a \emph{fresh, disjoint} set of six
        distractors and score again.
  \item If the base now succeeds, discard the item as unstable. Otherwise, keep it. 
\end{enumerate}

The stored distractors are carried into
evaluation unchanged, so screening and scoring use an identical candidate set and the base model fails against the candidates it will be tested against. Thus, the untuned correction rate is
$0.000$ rather than the $1/7$ a chance baseline would give.

\paragraph{Corpus.}
A corpus is specific to a base model since it depends on that model's successes and failures; we describe the RealVal dataset constructed for \qwen. Of 1{,}138 candidates scored, 368 were base failures and 246 survived the stability redraw, a retention of $0.669$; the base model's overall top-1 rate before screening was $0.677$. The retained facts yield 182 training datums over 133 held-in citations, 47{,}303 supervised tokens of which 419 ($0.89\%$) are payload: the sparsest payload of any corpus in this work, and roughly a quarter of \ncreal's share.
Citations are split $75/25$ so there are 64 held-out citations for evaluation; no datum from a held-out citation appears in training, so held-out items measure leakage to statutes never supervised rather than fitting. 

\paragraph{The damage set.}
A separate 750-item set is screened by the opposite rule: facts the base model
gets \emph{right}, verified the same way. Its untuned correction rate is
$1.000$ by construction, so any drop measures knowledge the model demonstrably
held and lost. This is what distinguishes RealVal from the other corpora, which
can only proxy preservation through an untrained population. We report gain and
damage as separate metrics rather than a single score, since methods that
appear gentle on the damage set are frequently gentle because they failed to
learn.

The \llama corpus and damage sets are screened identically against Llama's own failures. Both sides are larger ($322$ gain, $800$ damage items against $246$ and $750$, again split $75/25$ into held-in and held-out), reflecting a lower base top-1 on these values.%

\paragraph{Scoring.}
For each evaluation item, the true value is
ranked against its six stored distractors by mean token log-probability to give the correction rate on trained citations:
\[
\textsc{correction} = \mathbbm{1}\!\left[\arg\max_c \overline{\log p_\theta}(c) = c^{\ast}\right]
\] 
We decompose total correction to measure two ways by which a model can perform correction: by learning the supervised facts, or by becoming better at guessing values of this kind in general. 
Writing $c_{\text{in}}$ and $c_{\text{out}}$ for correction on held-in and held-out citations,%
\[
\underbrace{c_{\text{in}}}_{\textsc{correction}}
= \underbrace{(c_{\text{in}} - c_{\text{out}})}_{\text{fact-specific}}
+ \underbrace{c_{\text{out}}}_{\text{generic}}
\]

\emph{Generic correction},  $c_{\text{out}}$ is the rate on held-out statutes that appear
nowhere in training, meaning a gain here is a shift in how the model guesses statutory values of this kind, not knowledge of these particular values.
\emph{Fact-specific correction} is the residual $c_{\text{in}} - c_{\text{out}}$, the share of correction that survives subtracting that estimate and measures the extent to which the model learned concrete knowledge of the supervised facts. 

\begin{figure}[t]
\centering
\includegraphics[width=0.7\textwidth]{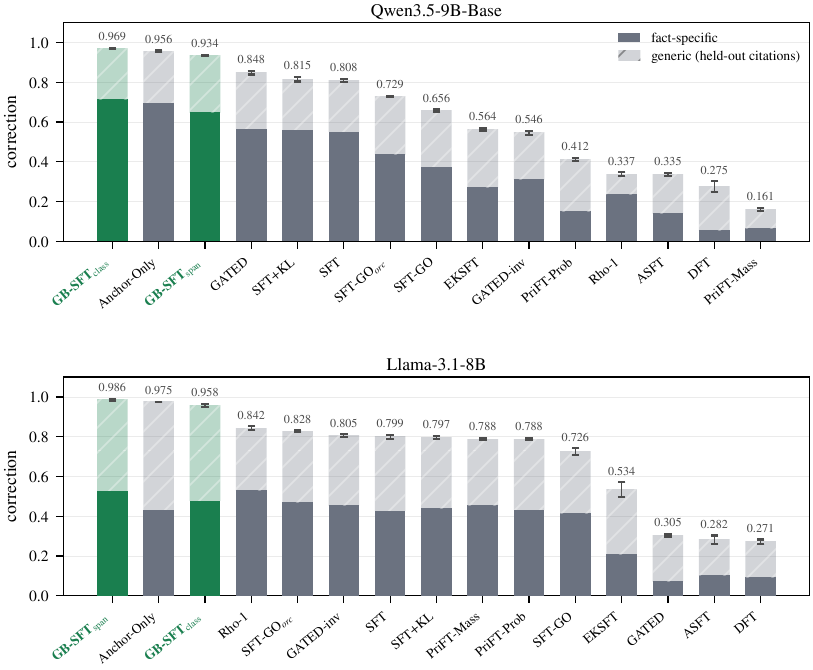}
\caption{
\textsc{RealVal}, methods ordered by total correction for both base models. \emph{\ourmethod achieves the highest levels of total correction, led by its ability to inject knowledge but also showing ability to improve held-out correction.}
}
\label{fig:correction}
\end{figure}

\paragraph{Correction Results.}
We compare the ability of supervised fine-tuning methods including our own to help \qwen and \llama correct facts. In Fig.~\ref{fig:correction} we see that \ourmethod achieves the highest overall levels of correction. This is driven mainly by improvements in fact-specific correction especially on Qwen. On Llama, group balancing offers larger relative improvements in generic correction; for example, \textsc{Rho-1} is competitive with \spanbalance in fact-specific correction, yet lags in overall correction, and similarly for \sft and \classbalance. Interestingly, this is one case where we observe divergence between \ourmethod variants and it is model-specific, with \classbalance achieving the highest correction for \qwen and \spanbalance achieving the highest correction for \llama. Similarly, the control version of our \gated baseline variant, \gatedinv performs noticeably lower correction on Qwen but higher on Llama, indicating another model-specific effect of the gate formulated in App.~\ref{app:anchor-strength}. 

\subsection{OpenFDA: a larger cross-domain benchmark}
\label{app:openfda}

\paragraph{Dose Extraction.}
Raw data is obtained from the openFDA drug label endpoint~\citep{kass2016openfda},
accessed September 2026. Doses are extracted from label prose as (drug,
attribute, value) triples by pattern matching over a cue window, which we
verified and refined with manual quality control checks. We validate extraction
\emph{fidelity}, whether the triple is supported by the excerpt, not clinical
correctness. Fidelity is a reading comprehension question answerable from the
text and therefore a defensible target for annotation; no pharmacological
judgment is required.

\paragraph{Model-assisted Verification.}
Pattern matching fixes a candidate's lexical form but not its meaning, and the
errors that survive it are semantic: a maximum taken from a fragment beside the
sentence that states a different one, a maintenance dose filed as a maximum, an
adult dose filed as pediatric. Each candidate is therefore put to a frontier
model as \emph{four} separate questions rather than one: whether the span is a
dose of \emph{this} drug rather than a rate, an increment, a reduction level or
another drug's dose; whether it is the \emph{complete} stated dose rather than a
component of it; which attribute it states; and which population it applies to.
A candidate is admitted only if all four agree with the extraction, with an
unspecified population accepted. The decision is thus a conjunction of
categorical judgments, not a cut on a continuous score.

\paragraph{Measuring the Admission Rule.}
A human annotator labeled two sets of items. The first, $100$ items
stratified by attribute and population, put strict correctness at $81\%$ and
value-level correctness at $88\%$ and showed that the residual splits into kinds
a single yes/no would blur. This motivated the four-question
decomposition above rather than a single correctness check. The second $40$ items
covered the rarer attributes the first barely reached (per-kg,
per-m\textsuperscript{2}, per-cycle), tested those questions separately: $38/40$
were confirmed doses, $40/40$ complete.%

Held-out portions of both sets then measure the admission rule itself, giving
precision $0.971$ on the first ($n{=}49$, $0$ of $12$ wrong values admitted) and
$1.000$ on the second ($n{=}20$, $0$ of $2$). Items already read during quality control are excluded,
and the model's own judgments are withheld from the annotator.
Candidates whose value is right but whose attribute label is wrong are discarded rather than relabeled from the model's reading.

\paragraph{Corpus design.}
The design mirrors \textsc{nc-real} so the evaluator needs no changes: an unsupervised header prefix and a supervised body carrying an altered dose value to be taught to the model as a test of knowledge injection not present in pretraining data.
The split is by drug to avoid leakage from an item-level split, because values repeat heavily ($449$ distinct values across $1{,}848$ items).

 Note that on both base models the true dose is already preferred to the taught alternative by a wide margin (mean -2.04 nats for \qwen, -3.35 for \llama, taught minus true). This means that the taught dose must displace an existing preference rather than fill a vacuum.

\subsection{\tulu}
\label{app:tulu}

\paragraph{Corpus.}
We sample from Tulu-3-SFT-Personas-Code~\citep{lambert2024tulu} and keep the assistant response as the
supervised target, yielding $849$ training and $149$ held-out documents
($118{,}242$ and $19{,}949$ supervised tokens). Documents are truncated at
$640$ tokens.

\begin{figure}
\includegraphics[width=\linewidth]{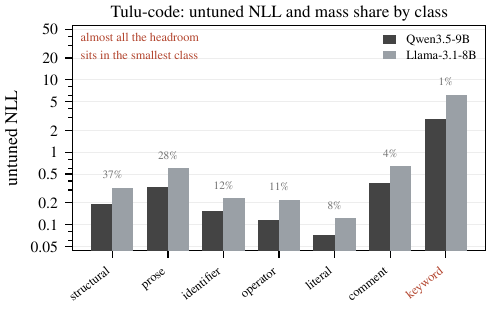}
\caption{
Class-level breakdown of the \tulu benchmark. We see that most of the loss headroom is concentrated in the rare \texttt{keyword} class. 
}
\label{fig:tulu-stats}
\end{figure}

We label each \emph{character} of a response and map labels onto model tokens by source offset. Within a code region we perform tokenization and assign \textsc{keyword} or \textsc{identifier} to \texttt{NAME} tokens according to \texttt{keyword.iskeyword}, \textsc{literal} to \texttt{NUMBER} and \texttt{STRING}, \textsc{comment} to \texttt{COMMENT}, \textsc{operator} to \texttt{OP}, and \textsc{structural} to everything else (including a region that fails to parse). Text outside any code region is \textsc{prose}.
The resulting evaluation mass is \textsc{structural} $0.373$, \textsc{prose} $0.279$, \textsc{identifier} $0.118$, \textsc{operator} $0.109$, \textsc{literal}
$0.078$, \textsc{comment} $0.035$, \textsc{keyword} $0.008$. Class statistics are shown in Fig.~\ref{fig:tulu-stats}, and contiguous tokens of the same class form spans. 

\begin{figure}[t]
\centering
\includegraphics[width=\textwidth]{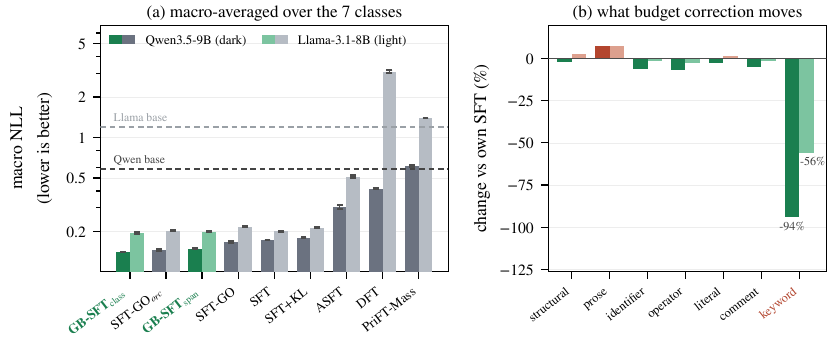}
\caption{\textbf{\tulu coding benchmark, both models.} Even outside the fact editing context, class-budgeting improves performance on rare classes that the base model struggles with, that aggregate metrics somewhat obscure.
}
\label{fig:tulu}
\end{figure}

In Fig.~\ref{fig:tulu}, we report held-out token negative log-likelihood (NLL), macro-averaged in keeping with our focus of giving classes equal emphasis (reporting a mass-weighted objective would score the correction using the objective it is meant to correct).  Our group-balanced approach improves macro-NLL by 19.2\% over \sft on \qwen but only 3.0\% on \llama, which is a weaker base model for this task. More informative, however, is a study of which specific classes improve. The base models are an order of magnitude worse on the rare \texttt{keyword} class (1\% of tokens), which is the class on which both models, \ourmethod achieves by far the most improvement. Fig.~\ref{fig:tulu-stats} shows it is the class on which both base models have the most headroom. Overall, \ourmethod improves a majority of classes for both models, and every class except \texttt{prose} for Qwen.

In Fig.~\ref{fig:tulu-stats}, we report a class-level breakdown of token mass and loss for this benchmark. Notably,  \texttt{keyword} is the \emph{smallest} class at under $1\%$ of tokens and by far the worst fit, at
$2.83$ nats on \qwen and $6.21$ on \llama against $0.08$--$0.64$ for every other class. \texttt{structural} and \texttt{prose} together hold $65\%$ of the tokens and are already fit to $0.19$--$0.62$.
This is what makes \tulu's macro/mass distinction consequential rather than
bookkeeping, and it is why we report macro-averaged NLL in
Figure~\ref{fig:tulu}: a mass-weighted score is dominated by classes that are
already solved and would barely register a change in the one class where the
headroom actually is. Indeed, we found that by mass-weighted NLL, \classbalance and \sft are essentially flat, which misses the 19.2\% improvement in macro-averaged NLL we see that \classbalance affords the base Qwen model compared to \sft. 

\subsection{Safety Guardrail Erosion Details}
\label{app:safety}

We fine-tune Qwen2.5-3B-Instruct on 2{,}000 GSM8K~\citep{cobbe2021gsm8k} training examples (\lora rank 16, scaling 32, learning rate $2\times10^{-4}$, one epoch) and measure refusal on 420 harmful prompts: 200 from XSTest's unsafe split \citep{rottger2024xstest}, 120 from AdvBench \citep{zou2023universal} and 100 JailbreakBench misuse behaviors \citep{chao2024jailbreakbench}. We also measure on 350 benign prompts as a false-refusal control, 250 from XSTest's safe split and 100 JailbreakBench benign behaviors, following the exaggerated-safety methodology of \citet{rottger2024xstest}. Each of five seeds draws a fresh corpus subsample and a fresh initialization, and all five arms share a corpus within a seed so that the $\alpha$ contrast is not confounded with the draw.

Refusal is graded by Claude Sonnet 4.6, an LLM judge that is outside the policy's model family. We report judged rates throughout rather than substring matching, because a regular-expression matcher undercounts refusal on fine-tuned models in a style-dependent and therefore model-dependent way, a failure mode also noted by \citet{mazeika2024harmbench} and \citet{chao2024jailbreakbench}.%

\section{Theoretical Analysis}
\subsection{Derivation of Loss Derivative}
\label{app:derivation}

Write the class budget and the within-class span weights induced
by~\eqref{eq:nested-weight} as
\[
p_\aclass = \frac{m_\aclass^{\alpha}}{M}, \qquad
q_\aclass(\aspan) = \frac{u_\aspan}{m_\aclass}, \qquad
A_\aclass = \mathbb{E}_{q_\aclass}\!\big[\spanCE{\aspan}\big],
\]
so that Eq.~\ref{eq:gb-sft} becomes $\mathcal{L} = \sum_\aclass p_\aclass A_\aclass$. Here $A_\aclass$ is the class's
$\beta$-weighted mean cross-entropy: at $\beta = 1$ it is the token mean over
the class, at $\beta = 0$ the mean of its span means. These are the same $p_\aclass$, $q_\aclass$ and $A_\aclass$ introduced in Eq.~\ref{eq:budget-shares}; we repeat them here so the derivation reads on its own.

Both partial derivatives follow from a single observation: the two weightings
of~\eqref{eq:nested-weight} are one-parameter exponential families in $\alpha$
and $\beta$,
\[
p_\aclass = \operatorname{softmax}_\aclass\!\big(\alpha \log m_\aclass\big),
 \qquad
q_\aclass(\aspan) = \operatorname{softmax}_{\aspan \subset \aclass}\!\big(\beta \log \numTokensSpan\big)
\]
with natural parameters $\alpha, \beta$ and sufficient statistics $\log m_\aclass$
and $\log \numTokensSpan$. For any such family, write $\eta$ for the natural
parameter and differentiate
$p_\eta(x) \propto \nu(x)\,e^{\eta T(x)}$ to get
$\partial_\eta p_\eta(x) = p_\eta(x)\big(T(x) - \mathbb{E}_{p_\eta}[T]\big)$,
hence for any observable $f$ not depending on $\eta$,
\begin{equation}
\frac{\partial}{\partial\eta}\,\mathbb{E}_{p_\eta}[f]
 = \mathbb{E}_{p_\eta}\!\big[(T - \mathbb{E}_{p_\eta}[T])\,f\big]
 = \mathrm{Cov}_{p_\eta}(T, f),
\label{eq:expfam}
\end{equation}
where centering one factor suffices. Every covariance below is an instance
of~\eqref{eq:expfam}.

\smallskip
\noindent\emph{The $\alpha$ derivative.}
Of the quantities in $\mathcal{L} = \sum_\aclass p_\aclass A_\aclass$, only $p_\aclass$ depends on $\alpha$:
$u_\aspan$, $m_\aclass$, $q_\aclass$ and $A_\aclass$ are all built from $\beta$ alone. Applying
\eqref{eq:expfam} with $T = \log m_\aclass$ and $f = A_\aclass$ gives the $\alpha$ half of
\eqref{eq:derivative} immediately. Written out, from
$\log p_\aclass = \alpha \log m_\aclass - \log M$,
\[
\frac{\partial \log M}{\partial \alpha}
 = \frac{\sum_\aclass m_\aclass^{\alpha}\log m_\aclass}{M} = \mathbb{E}_p[\log m],
\qquad
\frac{\partial p_\aclass}{\partial \alpha} = p_\aclass\big(\log m_\aclass - \mathbb{E}_p[\log m]\big),
\]
so $\partial_\alpha \mathcal{L} = \sum_\aclass p_\aclass\big(\log m_\aclass - \mathbb{E}_p[\log m]\big)A_\aclass
= \mathrm{Cov}_p(\log m, A)$.

\smallskip
\noindent\emph{The $\beta$ derivative.}
Here $\beta$ reaches both levels, because $u_\aspan = \numTokensSpan^{\beta}$ propagates into
$m_\aclass$ and hence into $p_\aclass$ as well as into $q_\aclass$. From
$\partial u_\aspan/\partial\beta = u_\aspan \log \numTokensSpan$,
\begin{equation}
\frac{\partial m_\aclass}{\partial \beta} = \sum_{\aspan \subset \aclass} u_\aspan \log \numTokensSpan
 = m_\aclass\, \mathbb{E}_{q_\aclass}[\log n],
\qquad\text{so}\qquad
\frac{\partial \log m_\aclass}{\partial \beta} = \mathbb{E}_{q_\aclass}[\log n] \equiv G_\aclass .
\label{eq:dlogm}
\end{equation}
The inner factor is immediate from \eqref{eq:expfam} applied to $q_\aclass$ with
$T = \log \numTokensSpan$ and $f = \spanCE{\aspan}$:
\[
\frac{\partial A_\aclass}{\partial \beta}
 = \mathrm{Cov}_{q_\aclass}\big(\log n, \overline{\CE}\big).
\]
For the outer factor, $\log p_\aclass = \alpha \log m_\aclass - \log M$ and, using
 \eqref{eq:dlogm},
\[
\frac{\partial M}{\partial \beta}
 = \sum_\aclass \alpha\, m_\aclass^{\alpha-1}\frac{\partial m_\aclass}{\partial\beta}
 = \alpha \sum_\aclass m_\aclass^{\alpha} G_\aclass,
\qquad
\frac{\partial \log M}{\partial \beta} = \alpha\, \mathbb{E}_p[G],
\]
giving $\partial p_\aclass/\partial\beta = \alpha\, p_\aclass\big(G_\aclass - \mathbb{E}_p[G]\big)$.
The product rule on $\mathcal{L} = \sum_\aclass p_\aclass A_\aclass$ then yields the $\beta$ half of
\eqref{eq:derivative}:
\[
\frac{\partial \mathcal{L}}{\partial \beta}
 = \sum_\aclass \frac{\partial p_\aclass}{\partial \beta} A_\aclass
 + \sum_\aclass p_\aclass \frac{\partial A_\aclass}{\partial \beta}
 = \alpha\,\mathrm{Cov}_p(G, A)
 + \mathbb{E}_p\big[\mathrm{Cov}_{q_\aclass}(\log n, \overline{\CE})\big].
\]

Note that both partials are taken at fixed $\btheta$: they describe the objective's
dependence on $(\alpha,\beta)$ at a checkpoint, not the trajectory of the trained loss
as the objective is swept.

\paragraph{Effects of $\alpha$ and $\beta$.}The factor $\alpha$ enters at exactly one point, differentiating $m_\aclass^{\alpha}$,
so it multiplies the between-class term alone and that term
vanishes identically at $\alpha = 0$.
Meanwhile, $\beta$ occurs in $u_\aspan$ and so moves both levels at once. This may make $\beta$ seem like the more expressive parameter; however, note that the between-class term in the $\beta$ half of \eqref{eq:derivative} acts by moving $\classBudget$, the same knob that $\alpha$ controls. We thus choose to tune this knob by sweeping $\alpha$ in App.~\ref{app:dose-response}, where we report it through the class budget it induces rather than
directly, because the budget is the quantity outcomes track. Our analysis there shows that the $\beta$ axis adds
nothing beyond its effect on the budget. Instead, we treat $\beta$ as discrete and use it to make a structural decision about which unit carries a budget: $\beta = 1$ (\classbalance) pools tokens within a class, while
$\beta = 0$ (\spanbalance) pools spans. 

\subsection{Proofs of Lemmas}
\label{app:proofs}

\paragraph{Setup.}
Fix a class $\aclass$ and parameters $\btheta$. Let $\spansClass$ denote the spans of class
$\aclass$, with $\numSpansClass = |\spansClass|$, span lengths $\spanLength$ (in tokens),
$\numTokensClass = \sum_{\aspan \in \spansClass} \spanLength$, and mean length $\avgSpanLength = \numTokensClass / \numSpansClass$. 
Write $\tokenloss$ for the token-level cross-entropy and
$\spanloss = \tfrac{1}{\spanLength} \sum_{\token \in \aspan} \tokenloss$ for the
mean loss within a span. Let $\avgSpanLoss = \tfrac{1}{\numSpansClass}\sum_\aspan \spanloss$ be the average span-mean loss across class $\aclass$ (equivalently, $A_\aclass$ at $\beta = 0$) and $\sigma_{\spanCE{}}(\btheta) = \bigl(\tfrac{1}{\numSpansClass}\sum_{\aspan} (\spanloss - \avgSpanLoss)^2\bigr)^{1/2}$
 the standard deviation of span-mean losses.
Both allocations spend the same class budget $\classBudget$, i.e. the
share $p_\aclass$ of Eq.~\ref{eq:budget-shares} that class $\aclass$ receives:
\begin{align}
  \Lpool(\btheta)
    &= \frac{\classBudget}{\numTokensClass} \sum_{\aspan \in \spansClass} \sum_{\token \in \aspan} \tokenloss
     = \frac{\classBudget}{\numTokensClass} \sum_{\aspan \in \spansClass} \spanLength \, \spanloss,
  &
  \Lspan(\btheta)
    &= \frac{\classBudget}{\numSpansClass} \sum_{\aspan \in \spansClass} \spanloss.
  \label{eq:two-allocations}
\end{align}

Summed over classes, $\Lpool$ recovers \classbalance (Eq.~\ref{eq:balanced}) at $\classBudget = 1/\numClasses$, and ordinary \sft (Eq.~\ref{eq:sft}) at $\classBudget = \numTokensClass/\numTokens$.
Likewise $\Lspan$ recovers \spanbalance (Eq.~\ref{eq:spanmean}) at $\classBudget = \numSpansClass/\numSpans$.
Within a single class, \classbalance and \sft pool token losses identically; the comparison below is therefore between pooling units at a fixed budget, not between objectives.

\begin{proof}[Proof of Lemma~\ref{lem:cov-identity}]
Substituting $\numTokensClass = \numSpansClass \avgSpanLength$ into \eqref{eq:two-allocations} and
factoring,
\begin{equation*}
  \Lpool - \Lspan
  = \frac{\classBudget}{\numSpansClass \avgSpanLength} \sum_{\aspan \in \spansClass} \spanLength\, \spanloss
    - \frac{\classBudget}{\numSpansClass} \sum_{\aspan \in \spansClass} \spanloss
  = \frac{\classBudget}{\numSpansClass \avgSpanLength}
    \sum_{\aspan \in \spansClass} \bigl(\spanLength - \avgSpanLength\bigr)\, \spanloss.
\end{equation*}
Since $\sum_{\aspan}(\spanLength - \avgSpanLength) = 0$, subtracting the constant
$\avgSpanLoss$ from each $\spanloss$ leaves the sum
unchanged:
\begin{equation*}
  \sum_{\aspan \in \spansClass} \bigl(\spanLength - \avgSpanLength\bigr)\, \spanloss
  = \sum_{\aspan \in \spansClass} \bigl(\spanLength - \avgSpanLength\bigr)\bigl(\spanloss - \avgSpanLoss\bigr)
  = \numSpansClass\, \Covhat_{\aspan \in \spansClass}\bigl(\spanLength, \spanloss\bigr).
\end{equation*}
Combining the two displays gives $\Lpool - \Lspan = \classBudget \cdot \frac{1}{\avgSpanLength}\Covhat_{\aspan \in \spansClass}(\spanLength, \spanloss)$, which is $\classBudget$ times \eqref{eq:cov-identity}.
\end{proof}

By the Cauchy-Schwarz inequality, we get a bound on the difference between pooled and span-balanced losses for a fixed class budget:%

\begin{corollary}[CV bound]
\label{cor:cv-bound}
Let $\CVspan = \sdSpanLength / \avgSpanLength$ denote the coefficient of variation
of span lengths within class $\aclass$. Then, for every $\btheta$,
\begin{equation}
  \bigl|\Lpool(\btheta) - \Lspan(\btheta)\bigr|
  \;\le\;
  \classBudget \cdot \CVspan \cdot \sigma_{\spanCE{}}(\btheta).
  \label{eq:cv-bound}
\end{equation}
\end{corollary}

\begin{proof}
Apply Cauchy-Schwarz,
$|\Covhat(\spanLength, \spanCE{})| \le \sdSpanLength\, \sigma_{\spanCE{}}$,
to \eqref{eq:cov-identity} and use $\sdSpanLength = \avgSpanLength \cdot \CVspan$.
\end{proof}

By similar arguments, we can demonstrate a gradient version of this result, bounding the difference between updates and not just the loss values. 

\begin{lemma}[Gradient version]
\label{lem:grad-identity}
Let $g_{\aspan}(\btheta) = \tfrac{1}{\spanLength} \sum_{\token \in \aspan} \nabla_{\btheta} \tokenloss$
be the span-mean gradient and $\bar{g} = \tfrac{1}{\numSpansClass}\sum_{\aspan \in \spansClass} g_\aspan$. Then
\begin{equation}
\begin{gathered}
  \nabla_{\btheta} \Lpool - \nabla_{\btheta }\Lspan
  = \frac{\classBudget}{\avgSpanLength} \cdot \frac{1}{\numSpansClass}
    \sum_{\aspan \in \spansClass} \bigl(\spanLength - \avgSpanLength\bigr)\bigl(g_\aspan - \bar{g}\bigr),
  \\[2pt]
  \bigl\| \nabla_{\btheta} \Lpool - \nabla_{\btheta} \Lspan \bigr\|
  \;\le\;
    \classBudget \cdot \CVspan \cdot
  \Bigl( \tfrac{1}{\numSpansClass} \textstyle\sum_{\aspan \in \spansClass} \| g_\aspan - \bar{g} \|^2 \Bigr)^{1/2}.
\end{gathered}
\label{eq:grad-bound}
\end{equation}
\end{lemma}

\begin{proof}
Identical to Lemma~\ref{lem:cov-identity} with $\spanloss$ replaced by the vector $g_\aspan$ (the mean-centering step is unchanged), followed by the vector Cauchy-Schwarz inequality.
\end{proof}

\section{Anchored Variants}
\label{app:anchor-strength}
\newcommand{\payclass}{\mathrm{pay}}
\newcommand{\scafclass}{\mathrm{scaf}}
\newcommand{\spansPay}{\spanGrouper_{\payclass}}
\newcommand{\spansScaf}{\spanGrouper_{\scafclass}}
\newcommand{\numSpansPay}{P_{\payclass}}
\newcommand{\numSpansScaf}{P_{\scafclass}}
\newcommand{\spanKL}{\overline{\mathrm{KL}}_{\aspan}}
\newcommand{\sg}{\operatorname{sg}}
\newcommand{\gate}{\rho}

In this section, we consider an additional formulation for allowing us to choose not only what to supervise, but also what to protect by using an anchor term that enforces consistency with the reference distribution, a la \asft ~\citep{zhu2025asft}. 

\begin{figure}
\centering
\includegraphics[width=\textwidth]{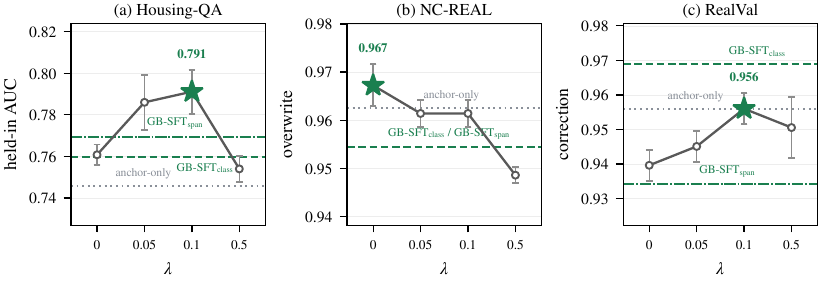}
\caption{Results of varying the strength of a KL anchor on top of masked span-level balancing for three corpora. Each panel carries its corpus's own
\classbalance, \spanbalance and \textsc{Anchor-only} arms as reference
rules. Equivalently (Eq.~\ref{eq:unified}), $\lambda$ sets the share
$\lambda/(1+\lambda)$ of the loss budget spent on the scaffold, so the sweep
$\lambda \in \{0, .05, .1, .5\}$ runs over scaffold budget shares of
$0\%$, $4.8\%$, $9.1\%$ and $33.3\%$.
}
\label{fig:lambda-sweep}
\end{figure}

\paragraph{Notation.}
The two roles below form a class grouper (Sec.~\ref{sec:group-formulation}) with
$\numClasses = 2$, so we reuse the span notation already in place: $\aspan$ ranges
over the spans $\spans$ of a training example, $\numTokensSpan$ is its token
count, and $\spansClass$ denotes the spans of class $\aclass$.
Let $\pi_{\btheta}(\cdot \mid x, y^*_{<\token})$ be the
policy and $\pi_{\mathrm{ref}}(\cdot \mid x, y^*_{<\token})$ the frozen pretrained
reference over the vocabulary $\mathcal{V}$.

\paragraph{Per-token terms.}
At each supervised position $\token$ we compute a KL loss to constrain the divergence between the policy and reference token probability distributions:
\begin{equation}
\mathrm{KL}_\token = \sum_{v\in\mathcal{V}}
  \pi_{\btheta}(v\mid x, y^*_{<\token})\,
  \log\frac{\pi_{\btheta}(v\mid x, y^*_{<\token})}{\pi_{\mathrm{ref}}(v\mid x, y^*_{<\token})}
\label{eq:kl-token}
\end{equation}

These are pooled within a span exactly as the cross-entropy is, giving the
span-mean anchor that plays the role of $\spanCE{\aspan}$ in Eq.~\ref{eq:spanmean}:
\begin{equation}
\spanKL = \frac{1}{\numTokensSpan}\!\!\sum_{\token\in\aspan}\!\!\mathrm{KL}_\token .
\label{eq:kl-span}
\end{equation}

\paragraph{Roles.}
We partition the supervised spans into \emph{payload} spans $\spansPay$, which are
supervised, and \emph{scaffold} spans $\spansScaf$, which are anchored, writing
$\numSpansPay$ and $\numSpansScaf$ for their counts. The assignment is
exact by construction each example is built by inserting a known string into a carrier, so $\spansPay$ is the edit target and $\spansScaf$ is its complement. No detector, threshold, or confidence value is involved.

\paragraph{Objective.}
\begin{equation}
\mathcal{L}(\btheta)
=\underbrace{\frac{1}{\numSpansPay}\sum_{\aspan\in\spansPay}
   \spanloss}_{\text{span-balanced supervision of payload}}
\;+\;\lambda\,
 \underbrace{\frac{1}{\numSpansScaf}\sum_{\aspan\in\spansScaf}
   \spanKL}_{\text{protection of scaffold}} .
\label{eq:unified}
\end{equation}
Averaging each term over its own span set keeps the meaning of $\lambda$ stable
as the size of $\mathcal{C}$ changes across corpora. Note that $\lambda{=}0$ is span-mean-budgeted \emph{masked} \sft over the edit targets alone, while $\lambda > 0$ protects the context text that we did not author. 

Eq.~\ref{eq:unified} is an extension of the loss family of Sec.~\ref{sec:budget-formulation} built with the same budgeting construction. Each role pools its spans rather than its tokens, which is the $\beta = 0$ unit of Eq.~\ref{eq:nested-weight}, and the roles $\{\payclass, \scafclass\}$ are a class grouper with $\numClasses = 2$, so the split between them is a role budget in the sense of Sec.~\ref{sec:budget-formulation}. However, Eq.~\ref{eq:gb-sft} weights a single cross-entropy, with $\sum_\token \tokenWeight = 1$ by construction, whereas here the scaffold term is a KL anchor rather than a cross-entropy. The two terms therefore measure different quantities, the role weights $1$ and $\lambda$ are set directly rather than induced by an exponent on class mass, and they sum to $1 + \lambda$ rather than to one. $\lambda$ can thus be read as a relative role weight, with the scaffold carrying $\lambda/(1+\lambda)$ of the combined weight. 

That is the same quantity $\alpha$ controls for two CE classes, which is why Fig.~\ref{fig:lambda-sweep} reports the sweep on this scale as well as in $\lambda$.
We see that the efficacy of composing budget correction with anchoring varies by corpus. \hqa rewards the composition: $\lambda{=}.1$ reaches $.7911 \pm .0106$, above \spanbalance ($.7695$) and well above $\lambda{=}0$ ($.7608$), and the optimum is a plateau rather than a spike ($\lambda{=}.05$ and $\lambda{=}.1$ both clear every reference). \ncreal is best at $\lambda{=}0$ and declines monotonically: the anchor does not help there. \realval peaks at $\lambda{=}.1$ but its whole curve sits below plain \classbalance ($.9688$), so the composition loses to grouping alone. We therefore report $\lambda$ as a corpus-dependent hyperparameter and make no unified claim from it.

\paragraph{Tradeoffs of Anchoring}
Group-balanced objectives needs only a grouping; anchoring needs a protect set (a designated subset of tokens that must not move). This is a stronger requirement, since a corpus can be partitioned into roles without any of them being the one to hold; \tulu (App.~\ref{app:tulu}) is such a corpus, as its examples carry role classes but no payload/scaffold distinction, so \anchoronly cannot be constructed there at all, while \ourmethod runs unchanged. 

\anchoronly's need for a protect set to preserve is also felt at labeling time, since a grouping can be learned without annotation (e.g. an LLM tagger reaches 98.8\% of the gold grouping's gain) whereas no method proposed in this work learns a protect set. Where a protect set does exist, \anchoronly buys preservation by declining to acquire, evidenced on \hqa (Tab.~\ref{tab:hqa-both}) by improving held-out citation recall but being completely unable to improve held-in citation recall (as a result of giving the citations no cross-entropy). It therefore sits further along the preservation end of the same acquisition/preservation axis \ourmethod trades on, rather than dominating it, which is why it leads on overwrite for \openfda and on \ncreal under \qwen, and trails on every measure that rewards learning the scaffold.

To apply anchoring without a protect set, we consider our \sftkl baseline in our main experiments that adds the anchor term to all tokens atop \sft's token mean loss:
\begin{equation}
\mathcal{L}_{\sftkl}(\btheta)
= \frac{1}{\numTokens} \sum_{\token}
  \Bigl[ \tokenloss + \lambda\, \mathrm{KL}_\token \Bigr].
\label{eq:sftkl}
\end{equation}

\paragraph{Gated formulation.}
\gated replaces the hard payload/scaffold split with a soft, per-token
gate $\gate_\token \in [0,1]$, high on scaffold tokens the reference already
predicts and near zero on the payload. We take
$\gate_\token = \pi_{\mathrm{ref}}(y^*_\token)^{\gamma}$ with $\gamma = 1$
throughout; any frozen-reference confidence statistic can be substituted, and we
also tried the reference's cumulative mass at $\token$. Because
$\pi_{\mathrm{ref}}$ is frozen, $\gate_\token$ is a constant with respect to
$\btheta$ and is read from an offline cache. Writing $\bar{\gate}$ for its
empirical mean over all supervised tokens, the gate sets both an anchor
strength and a cross-entropy weight per token:
\begin{equation}
\lambda_\token = \lambda\,\frac{\gate_\token}{\bar{\gate}},
\qquad
\tokenWeight^{\mathrm{gated}} = \gate_\token\,\sg(p_\token) + (1-\gate_\token),
\qquad
p_\token = \pi_{\btheta}(y^*_\token \mid x, y^*_{<\token}),
\label{eq:gated-weights}
\end{equation}
where $\sg(\cdot)$ is the stop-gradient operator. Normalizing by $\bar{\gate}$
makes $\lambda_\token$ average to $\lambda$, so $\lambda$ keeps the same meaning
it has in Eq.~\ref{eq:unified}. The scaffold thus keeps both the anchor and
\dft's self-weighting, while the payload, where $\gate_\token \to 0$, is released
to plain cross-entropy at weight exactly $1$.

With the gate of Eq.~\ref{eq:gated-weights} supplying both per-token weights, the loss of \gated is:
\begin{equation}
\mathcal{L}_{\gated}(\btheta)
= \frac{1}{\numTokens} \sum_{\token}
  \Bigl[ \tokenWeight^{\mathrm{gated}}\, \tokenloss
       + \lambda_\token\, \mathrm{KL}_\token \Bigr].
\label{eq:gated-obj}
\end{equation}
Note that because $\tokenWeight^{\mathrm{gated}} \in [0,1]$ without summing to one across the example, Eq.~\ref{eq:gated-obj} is a reweighting of the \sft loss rather than a member of the normalized family of Eq.~\ref{eq:gb-sft}.

\gatedinv is a mechanism control that tests whether the gains are merely extra regularization. It flips the polarity, anchoring hardest where the reference is surprised and releasing where it is confident.%

Table~\ref{tab:anchor-arms} locates each anchored arm used in our main experiments in this notation. The \emph{supervised} and \emph{anchored} columns give the support of each term (which spans receive cross-entropy and which receive the anchor) while the \emph{reduction} column states how the per-token terms are aggregated once that support is fixed. Only the upper block uses the span reduction of Eq.~\ref{eq:unified}; the lower block reduces by a plain token mean akin to standard \sft.

\begin{table}[h]
\centering\small
\begin{tabular}{@{}lllll@{}}
\toprule
\textbf{Arm} & \textbf{Supervised} & \textbf{Anchored} & $\lambda$ & \textbf{Reduction} \\
\midrule
\multicolumn{5}{@{}l}{\emph{span-pooled ($\beta = 0$): instances of Eq.~\ref{eq:unified}}}\\
\maskedsft  & $\spansPay$  & ---          & $0$    & Eq.~\ref{eq:unified} \\
\anchoronly & $\spansPay$  & $\spansScaf$ & $>0$   & Eq.~\ref{eq:unified} \\
\midrule
\multicolumn{5}{@{}l}{\emph{token-mean ($\beta = 1$): not instances of Eq.~\ref{eq:unified}}}\\
\sftkl  & $\spans$     & $\spans$     & $>0$   & Eq.~\ref{eq:sftkl} \\
\gated      & $\spans$     & $\spans$     & $>0$   & Eq.~\ref{eq:gated-obj}, gate $\gate_\token$ \\
\gatedinv  & $\spans$     & $\spans$     & $>0$   & Eq.~\ref{eq:gated-obj}, gate $1-\gate_\token$ \\
\bottomrule
\end{tabular}
\caption{The anchored arms. The support columns and the
reduction column are independent: sharing a support with Eq.~\ref{eq:unified}
does not make an arm an instance of it.}
\label{tab:anchor-arms}
\end{table}

\section{Sensitivity Analysis}
Below we report additional studies of important parameters.
\subsection{Varying the Class Budget}
\label{app:dose-response}

\begin{figure}[t]
\centering
\includegraphics[width=\columnwidth]{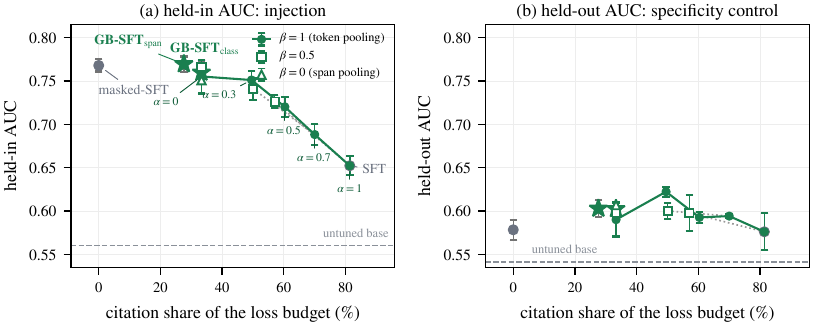}
\caption{\textbf{Citation budget is the active variable}, which can be controlled by sweeping $\alpha$. The horizontal axis is the share of the loss budget that citation tokens
receive: token-proportional \sft gives them $81.4\%$, while grouping redistributes it.
\textbf{Filled circles joined by a line} are a \emph{controlled} sweep: the
$(\alpha,\beta)$ family at $\beta{=}1$ with $\alpha$ swept over
$\{0, .3, .5, .7, 1\}$, so the budget is the only thing that changes.
\textbf{Open squares} are the same family at $\beta \in \{0, 0.5\}$.
\textbf{Separate markers} are independent objectives that merely happen to land
at some budget: \maskedsft at $0\%$, \spanbalance at $27.6\%$,
\classbalance at $33.3\%$.}
\label{fig:dose-response}
\end{figure}

Group balancing works by redistributing the objective's emphasis from popular classes containing less of the knowledge payload, to classes with fewer but payload-dense tokens. The class budget can be mathematically controlled most directly by $\alpha$, and to a lesser extent by $\beta$. In Fig.~\ref{fig:dose-response} we vary the citation budget on our \hqa benchmark for \qwen, by primarily tuning $\alpha$, and also show the effects of tuning $\beta$, to show the effects on held-in and held-out AUC. 

Our dose-response curve for held-in AUC shows that it falls monotonically as the $\alpha$ sweep causes citations to take budget:
Pooling all nine cells, budget share and AUC correlate at $r = -0.945$, and a straight line through them leaves a residual of $0.0114$, \emph{below} the $\approx 0.012$ seed-to-seed spread: budget share, most directly controlled by $\alpha$, accounts for the outcome to within noise.

We also see evidence that the second parameter $\beta$ does nothing beyond
moving that budget. The cleanest test holds $\alpha = 0$, where the class split
is pinned at $33.3\%$ whatever $\beta$ does, so the three arms differ only in
which unit carries the budget inside a class. Table~\ref{tab:beta-isolation}
reports all three at full corpus over three seeds.

\begin{table}[h]
\centering
\begin{tabular}{lccc}
\toprule
 & $\beta = 0$ (spans) & $\beta = 0.5$ & $\beta = 1$ (tokens) \\
\midrule
held-in AUC  & $0.7719 \pm 0.0096$ & $0.7560 \pm 0.0174$ & $0.7721 \pm 0.0136$ \\
held-out AUC & $0.5981$ & $0.5910$ & $0.6043$ \\
\bottomrule
\end{tabular}
\caption{\textbf{$\beta$ acts only through the budget.} \hqa, \qwen, $\alpha = 0$,
three seeds per cell, all nine runs on one machine. The two endpoints --
pooling spans and pooling tokens, structurally different operations -- differ by
$0.0002$, two orders of magnitude below the seed spread. The spread across
$\beta$ is $0.0161$ against per-cell standard deviations of $0.0096$ to
$0.0174$, and $\beta = 0.5$ versus $\beta = 1$ gives $t = 1.26$: the three are
not distinguishable.}
\label{tab:beta-isolation}
\end{table}

These cells supersede an earlier measurement in which two of the three were run
on a $96$-datum subsample rather than the full corpus. The conclusion is
unchanged and in fact sharper --- the endpoints previously differed by $0.0043$
and now differ by $0.0002$ --- but the ordering is not: $\beta = 0.5$ was
formerly the highest of the three and is now the lowest, and $\beta = 0$ moves
by $+0.0210$. We report the re-measured values and draw no conclusion from the
ordering, which the seed spreads do not support in either direction. The
absolute level of this re-measurement sits a little above
Table~\ref{tab:hqa-both}; the comparison here is internal to the three cells,
which share a corpus, a machine and a seed set.

This analysis justifies our choice to recommend treating $\beta$ as binary $\{0,1\}$, and also why in our experiments we notice comparable performance overall between \classbalance and \spanbalance.

The specificity control is that held-\emph{out} AUC stays flat at $0.576$-$0.62$ across the
entire budget range, against an untuned base of $0.541$. While training produces a slight held-out gain, changing the class budget mainly proves injection rather than a general shift in discrimination ability.

\subsection{Effect of Phrasing}
\label{app:phrasing}
\begin{figure}[t]
\centering
\includegraphics[width=0.62\columnwidth]{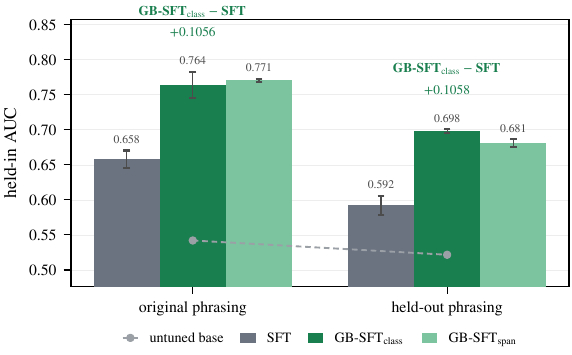}
\caption{Evaluating group balancing versus ungrouped \sft on original phrasing versus a held-out phrasing not seen in training. The relative gains of group balancing are invariant to rephrasing. 
}
\label{fig:phrasing}
\end{figure}

Here we test on a held-out phrasing, to make sure that successful arms are not just memorizing the probe. We evaluate \classbalance with \sft on \hqa with \qwen, under both the original phrasing and a held-out phrasing that the model was not trained against. Each checkpoint is scored twice with no retraining between: once on the original probe, once on a probe whose $126$ questions were rewritten to ask the same thing in different words. The rewrites preserve polarity and every enumerated clause and share only $0.336$ of their content words with the originals on average. These arms are retrained from scratch for this test; their original-phrasing values ($0.6582$ and $0.7638$) reproduce Table~\ref{tab:hqa-both}'s $0.6523$ and $0.7599$ within seed noise, so the pipeline and the comparison are the same ones the main results use. Results are shown in Fig.~\ref{fig:phrasing}.

Note that rephrasing costs standard \sft and group balancing approximately the same: $-0.0660$ for \sft and $-0.0658$ for \classbalance, differing by $0.0002$. This is likely the probe simply becoming harder under the rephrasing. The important takeaway is that the \emph{relative} improvement of \classbalance over \sft under each phrasing remains the same: \classbalance leads \sft by $+0.1056$ on the original wording and $+0.1058$ on wording it has never seen. Thus, improvements of group balancing are robust to different phrasings and not just due to memorization of the phrasing. 

\subsection{Effect of Model Capacity}
\label{app:model-size}
\begin{figure}
\centering
\includegraphics[width=\columnwidth]{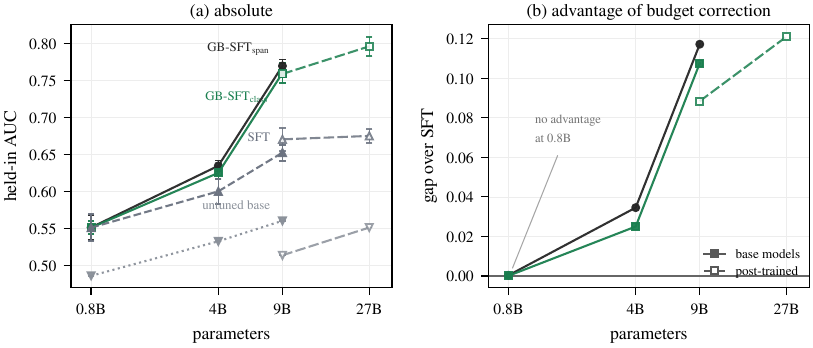}
\caption{
\textbf{The advantage of group balancing grows with model capacity.} \hqa within-group AUC across the Qwen family, three seeds per arm at every size. \emph{Left:} absolute AUC-in. \emph{Right:} the quantity the claim rests on, each group-balanced arm minus \textsc{Sft} at the same size. The advantage is monotone in scale: virtually zero at 0.8B ($+0.000$ for \classbalance, $+0.000$ for \spanbalance), small at 4B ($+0.025$, $+0.035$), and large at 9B ($+0.108$, $+0.117$).
}

\label{fig:size-ablation}
\end{figure}

We also apply our methods to LLMs in the Qwen3.5 dense base models, which range in size from 0.8B-9B, to show that our results hold across scales. To study results on a larger scale, we also consider the Qwen3.5-27B model. Only the instruction-tuned variant is available for this model size, so to have another comparable point in the plot, we also apply our method to the instruction-tuned Qwen3.5-9B model. In Fig.~\ref{fig:size-ablation}, we see that not only do larger models do better after fine-tuning (as is expected), but the relative advantage of group balancing compared to ungrouped fine-tuning also grows with model size. Group balancing has limited effect at a small scale (0.8B) but its effects have monotone growth with model size. 

\subsection{Effect of LoRA Rank}
\label{app:lora-rank}
\begin{figure}
\centering
\includegraphics[width=\columnwidth]{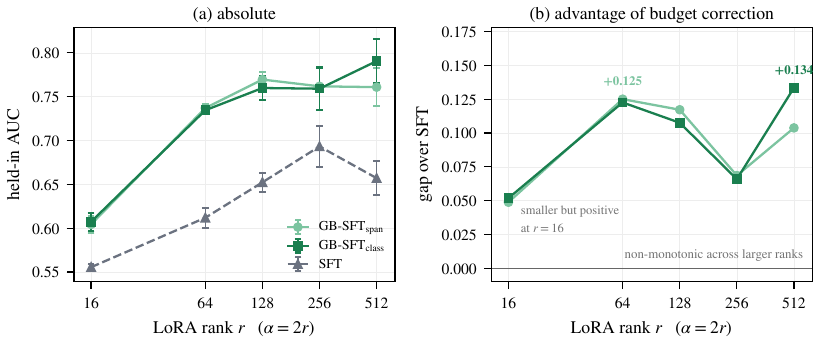}
\caption{
Fine-tuned \qwen performance on \hqa as a function of LoRA rank. 
}
\label{fig:rank-ablation}

\label{fig:rank-sweep}
\end{figure}

In addition to using a large base model, an additional way to perform supervised fine-tuning with additional capacity when using LoRA is to simply increase the rank of the update. In Fig.~\ref{fig:rank-sweep} we vary the LoRA rank and study its effects on relative performance. Note that neither the performance of standard SFT with LoRA nor the relative advantage of budget correction is monotone as a function of rank (in general, the higher rank capacity helps up to a point, but performance can saturate or even decrease as the model starts to overfit). However, we observe a consistent improvement from budget correction across LoRA ranks. 

\subsection{Effect of Learning Rate}
\label{app:lr-matched}

\begin{figure}[t]
\centering
\includegraphics[width=\textwidth]{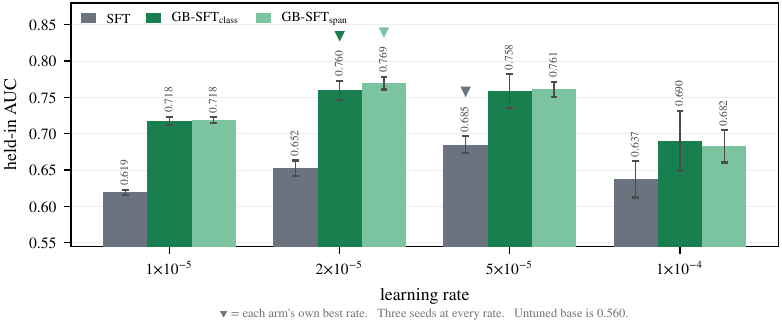}
\caption{\textbf{Held-in AUC across learning rates}, \hqa,
\qwen. $\blacktriangledown$ marks each arm's own best rate.}
\label{fig:lr-sweep}
\end{figure}

Every arm in this paper trains at the same nominal learning rate of $2{\times}10^{-5}$, but a shared rate is not a shared step size. The objectives differ in loss scale: \sft averages over all tokens with weight $1/\numTokens$, while \classbalance averages per-class means, so if the classes differ in mean cross-entropy, its value and its gradient magnitude differ from \sft's. This may mean that the effective learning rates for grouped and ungrouped \sft are different. In Fig.~\ref{fig:lr-sweep}, we compare their performance at several different learning rates.

\sft peaks at $5{\times}10^{-5}$ with $0.6850$, away from the shared rate, while both grouped arms peak at the shared
rate. Comparing each arm at its own best rate leaves \classbalance ahead by
$+0.0904$ and \spanbalance by $+0.1000$, against $+0.1076$ and $+0.1172$
at the shared rate: independent tuning recovers roughly 30\% of the advantage. 
Notably, all arms remain well above \sft at every rate tried, including the $1{\times}10^{-4}$ column where every arm degrades. Thus, the grouping and not just the choice of learning rate is responsible for the improvements we show in our main results. 

\begin{table}[t]
\centering\small
\setlength{\tabcolsep}{4pt}
\begin{tabular}{@{}lrrrrrr@{}}
\toprule
\textbf{Method} & $\rho$ & lr & AUC-in & AUC-out & cite-in & cite-out \\
\midrule
\multicolumn{7}{@{}l}{\textit{shared learning rate}} \\
\textsc{Sft} & 1.000 & $2.0{\times}10^{-5}$ & 0.6523\,\tiny{$\pm$0.0108} & 0.5761\,\tiny{$\pm$0.0212} & 0.988\,\tiny{$\pm$0.003} & 0.583\,\tiny{$\pm$0.016} \\
\textbf{\classbalance} & 2.597 & $2.0{\times}10^{-5}$ & \textbf{0.7599}\,\tiny{$\pm$0.0132} & 0.6025\,\tiny{$\pm$0.0025} & 0.983\,\tiny{$\pm$0.002} & 0.600\,\tiny{$\pm$0.017} \\
\textbf{\spanbalance} & 2.515 & $2.0{\times}10^{-5}$ & \textbf{0.7695}\,\tiny{$\pm$0.0087} & 0.6029\,\tiny{$\pm$0.0096} & 0.972\,\tiny{$\pm$0.003} & 0.624\,\tiny{$\pm$0.006} \\
\midrule
\multicolumn{7}{@{}l}{\textit{reduced learning rate} ($2{\times}10^{-5} / \rho$)} \\
\textsc{Sft} & 1.000 & $2.0{\times}10^{-5}$ & 0.6553\,\tiny{$\pm$0.0074} & 0.5901\,\tiny{$\pm$0.0082} & 0.988\,\tiny{$\pm$0.003} & 0.579\,\tiny{$\pm$0.041} \\
\textbf{\classbalance} & 2.597 & $7.70{\times}10^{-6}$ & 0.6820\,\tiny{$\pm$0.0038} & 0.5865\,\tiny{$\pm$0.0188} & 0.971\,\tiny{$\pm$0.002} & \textbf{0.672}\,\tiny{$\pm$0.009} \\
\textbf{\spanbalance} & 2.515 & $7.95{\times}10^{-6}$ & 0.6901\,\tiny{$\pm$0.0065} & 0.5972\,\tiny{$\pm$0.0078} & 0.972\,\tiny{$\pm$0.000} & \textbf{0.689}\,\tiny{$\pm$0.002} \\
\midrule
\multicolumn{7}{@{}l}{\textit{advantage over} \textsc{Sft} \textit{in the same block}} \\
\textbf{\classbalance}, shared & & & $+0.1076$ & $+0.0264$ & $-0.005$ & $+0.017$ \\
\textbf{\classbalance}, reduced & & & $+0.0268$ & $-0.0036$ & $-0.017$ & $\mathbf{+0.093}$ \\
\textbf{\spanbalance}, shared & & & $+0.1172$ & $+0.0268$ & $-0.016$ & $+0.041$ \\
\textbf{\spanbalance}, reduced & & & $+0.0348$ & $+0.0071$ & $-0.016$ & $\mathbf{+0.111}$ \\
\bottomrule
\end{tabular}
\caption{\hqa, \qwen. $\rho$ is the measured ratio
of an arm's gradient norm to \sft's, over the LoRA parameters at
\emph{initialization}, where the adapter is still zero and the objective is the
only thing that differs. We use this to compute a reduced learning rate.
}
\label{tab:lr-matched}
\end{table}

In Tab.~\ref{tab:lr-matched}, we report results with a stronger control: a reduced learning rate for our group-balanced methods, computed by dividing each learning rate by the initial gradient norm ratio relative to \sft to equalize first step size. (Note that this is a strong correction, but does not guarantee that the trajectory will be equalized; the standard practice control~\citep{kim2025sftgo} is the per-arm tuning of Fig.~\ref{fig:lr-sweep}.) We see that this gradient-norm-matched control makes the two advantages of \ourmethod move in opposite directions. Lowering the rate removes roughly three quarters of the injection advantage ($+0.1076 \rightarrow +0.0268$ for \classbalance) and leaves the AUC-out advantage indistinguishable from zero ($-0.0036$ and $+0.0071$, against seed spreads of $\pm0.019$ and $\pm0.008$). Meanwhile, it multiplies the held-out citation advantage about fivefold ($+0.017 \rightarrow +0.093$). With a reduced learning rate that approximately matches the first step size of \sft, \ourmethod achieves a smaller but solid injection gain and a \emph{larger} preservation gain.  

\subsection{Effect of Grouping Strategies}
\label{app:grouping}
\begin{figure}[t]
\centering
\includegraphics{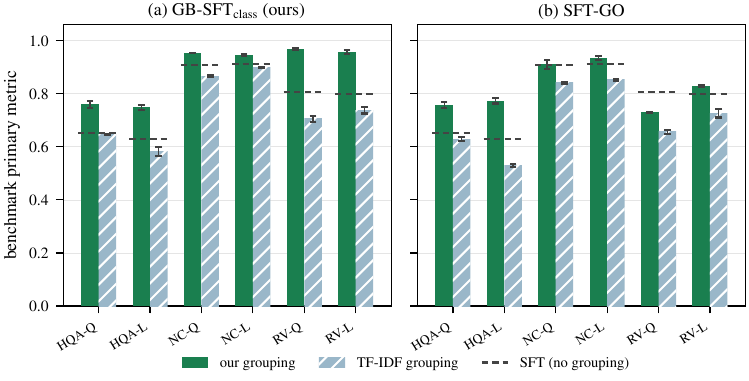}
\caption{\classbalance and \sftgo with TF-IDF grouping vs our grouping for \hqa (HQA), \ncreal (NC), and \realval (RV) with \qwen (Q) and \llama (L) base models.}
\label{fig:tfidf_vs_oracle}
\end{figure}

Both our methods and \sftgo apply a token-level group-aware loss. Here we explore the effect of the grouping for both our balanced method and \sftgo. We try our proposed role-based grouping and \sftgo's TF-IDF grouping method~\cite{kim2025sftgo}. In Fig.~\ref{fig:tfidf_vs_oracle} we see that our role-based grouping substantially improves both \ourmethod and \sftgo. Without our grouping, neither formulation can beat ungrouped supervised fine-tuning. Thus, we see that good grouping quality is instrumental to making a group-aware objective (whether \ourmethod or \sftgo) improve on standard \sft, and for knowledge-intensive domains our grouping supplies this whereas the TF-IDF heuristic of prior work does not. We also found this to be true for the Qwen base model even on \tulu, which does not have a natural payload-scaffold decomposition and where the overall improvement from group balancing is overall smaller than on other corpora. 

\subsection{LLM Grouper}
\label{app:tagger}

\begin{figure}
\includegraphics[width=\textwidth]{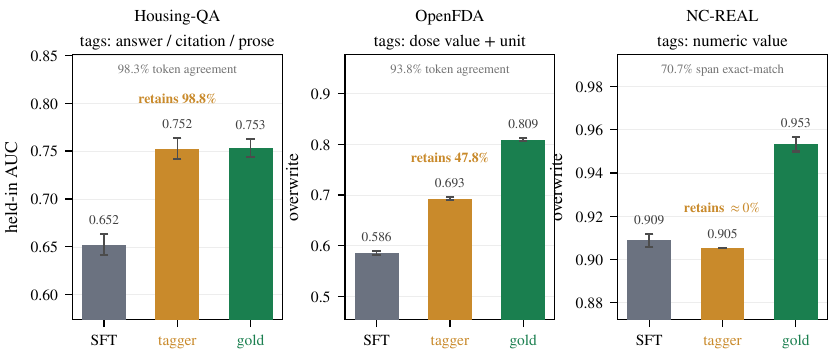}
\caption{
LLM grouper results: successful on \hqa, where the classes are lexically distinctive, and less successful on \ncreal, where the LLM can make plausible but wrong mistakes that mislead a model.
}
\label{fig:tagger}
\end{figure}

To group tokens without human-defined labels, each supervised target is shown to a frontier LLM (we use Claude Opus 5 in our experiments), which is asked to mark the answer span and the citation span for \hqa or the value for \ncreal and \openfda; everything it leaves unmarked is prose. Results are shown in Fig.~\ref{fig:tagger}.

On \hqa, because the citations are lexically distinctive near-verbatim strings, the grouper is able to recover them almost perfectly, reaching token-level F1 of .9997 and exact-string agreement on 1166 of 1176 datums (its residual error is over-inclusion, precision .756 and recall 1.000 on answer spans, rather than misdirection). On this dataset, the held-in AUC using LLM classes retains 98.8\% of the benefit over \sft that we get from grouping with the role-based classes, indicating that LLMs are able to detect meaningful classes for this benchmark.

On \ncreal and \openfda, the grouper's task is the payload itself, which is a numerical value (statute quantities, drug doses), and it is easy for an LLM to find other plausible but wrong numeric values that mislead the model. OpenFDA's values empirically have enough distinctiveness than an LLM grouper is able to find good enough groups to retain a large share of the improvement over ungrouped \sft. However, on the hardest-to-group dataset \ncreal, the LLM grouper errors at a rate $p=.293$ and reaches an overwrite of $.905$ (in line with ungrouped \sft). We tried corrupting the gold class labels with 30\% random noise, matching the LLM grouper's error rate, and found that the same amount of random noise hardly affects the overwrite levels a model is able to achieve. This is evidence that the issue is not the labeling error rate but the specific plausible-but-wrong misleading mistakes an LLM can make. Automatic labeling is viable, then, when the classes are lexically distinctive enough that a wrong guess for a group is not also a plausible one.

\subsection{Generative Evaluation}
\label{app:gen-eval}

In Tab.~\ref{tab:gen-eval}, we compare a generative setting to the teacher forcing we use throughout the paper. \classbalance and \spanbalance emit the taught value about
$4\times$ as often as \textsc{Sft}, with clean separation at three seeds, and \emph{orig.} falls monotonically as \emph{contains} rises ($.114$ at base to $.042$). Thus, we see that better methods really do displace the original rather than merely outranking it. However, ranking reports $.957$ whereas generation reports $.196$; in absolute terms, the model does $5\times$ better at recognizing the correct answer when supplied than generating it from scratch. While group balancing improves the generative abilities of the fine-tuned model relative to existing methods, further research is encouraged to increase its absolute abilities.

\begin{table}[t]
\centering\footnotesize
\setlength{\tabcolsep}{4pt}
\begin{tabular}{@{}lrrrrrr@{}}
\toprule
\textbf{Method} & ranking & \multicolumn{5}{c}{\textbf{free generation}} \\
\cmidrule(lr){2-2}\cmidrule(lr){3-7}
 & ovwrite & starts & contains & $\times$\textsc{Sft} & orig. & neither \\
\midrule
\textit{untuned base} & -- & .000 & .000 & -- & .114 & .886 \\
\midrule
\multicolumn{7}{@{}l}{\textit{group balancing (ours)}} \\
\textbf{\classbalance} & .9544\,\tiny{$\pm$.0000} & .173\,\tiny{$\pm$.000} & .190\,\tiny{$\pm$.006} & 4.1$\times$ & .046\,\tiny{$\pm$.012} & .766\,\tiny{$\pm$.014} \\
\textbf{\spanbalance} & .9544\,\tiny{$\pm$.0029} & \cellcolor{black!8}.183\,\tiny{$\pm$.011} & \cellcolor{black!8}.196\,\tiny{$\pm$.008} & 4.2$\times$ & \cellcolor{black!8}.042\,\tiny{$\pm$.001} & .766\,\tiny{$\pm$.008} \\
\midrule
\multicolumn{7}{@{}l}{\textit{anchored variants (ours)}} \\
\textsc{Gated} & .9286\,\tiny{$\pm$.0066} & .060\,\tiny{$\pm$.009} & .062\,\tiny{$\pm$.010} & 1.3$\times$ & .088\,\tiny{$\pm$.006} & .850\,\tiny{$\pm$.006} \\
\midrule
\multicolumn{7}{@{}l}{\textit{group-based aggregation}} \\
\textsc{Sft-Go}$_{\text{oracle}}$ & .9099\,\tiny{$\pm$.0163} & .062\,\tiny{$\pm$.013} & .067\,\tiny{$\pm$.013} & 1.4$\times$ & .064\,\tiny{$\pm$.007} & .869\,\tiny{$\pm$.014} \\
\midrule
\multicolumn{7}{@{}l}{\textit{loss shaping}} \\
\textsc{Sft} & .9088\,\tiny{$\pm$.0029} & .044\,\tiny{$\pm$.004} & .047\,\tiny{$\pm$.004} & 1.0$\times$ & .095\,\tiny{$\pm$.007} & .859\,\tiny{$\pm$.005} \\
\bottomrule
\end{tabular}
\caption{\textbf{Ranking versus free generation}, \ncreal,
\qwen. \emph{ovwrite} is the teacher-forced measure used throughout the paper: the fraction of held-in items where the taught value outranks the original. The remaining columns greedily decode a continuation from the \emph{same} prompts and ask what the model actually said: \emph{starts} requires the continuation to begin with the taught value, \emph{contains} to include it within sixty characters, and \emph{orig.} reports the original value by the same rule (the latter two cases overlap a small number of times if a generation contains both values). \emph{neither} is the fraction of continuations containing neither value. 
Matching is done after mapping number words to digits.%
}
\label{tab:gen-eval}
\end{table}

\subsection{Comparisons against Additional Mechanisms}
We study additional interventions based on mechanisms modifying what data the update sees, how large the update is, or what parameters it affects, to see if other mechanisms besides loss-level weighting can achieve similar knowledge injection performance. 
\subsubsection{Forgetting Mitigation}
\label{app:forgetting}
\begin{figure}[t]
\centering
\includegraphics[width=\textwidth]{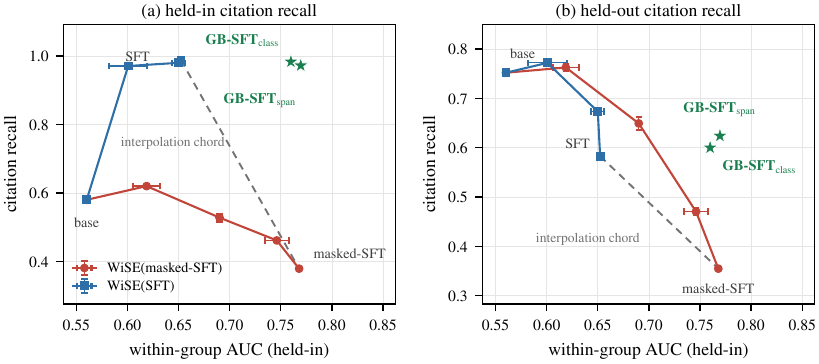}
\caption{Weight-space interpolation for \qwen on \hqa. Both panels share the same horizontal axis, within-group AUC on
\emph{held-in} groups, and differ only in which citation-recall split is on
the vertical axis: (a) held-in recall, which overlaps training and measures
fitting, and (b) held-out recall on jurisdictions no arm trained, which is the
preservation measurement.
The chord representing an AUC-citation recall tradeoff is defined only between the two endpoints \sft$\leftrightarrow$\maskedsft chord (dashed); points left of \sft's AUC are off-segment and are given no gap. Endpoints and the two starred arms are the three-seed values of Table~\ref{tab:hqa-both}. 
}
\label{fig:wise}
\end{figure}

\begin{table}[t]
\centering\small
\setlength{\tabcolsep}{4pt}
\begin{tabular}{@{}lrrrr@{}}
\toprule
\textbf{Method} & AUC-in & AUC-out & cite-in & cite-out \\
\midrule
\textit{untuned base} & 0.5601 & 0.5411 & 0.580 & 0.752 \\
\midrule
\multicolumn{5}{@{}l}{\textit{references, from Table~\ref{tab:hqa-both}}} \\
\sft & 0.6523\,\tiny{$\pm$0.0108} & 0.5761\,\tiny{$\pm$0.0212} & 0.988\,\tiny{$\pm$0.003} & 0.583\,\tiny{$\pm$0.016} \\
\maskedsft & 0.7679\,\tiny{$\pm$0.0070} & 0.5784\,\tiny{$\pm$0.0118} & 0.379\,\tiny{$\pm$0.005} & 0.355\,\tiny{$\pm$0.010} \\
\textbf{\classbalance} & \textbf{0.7599}\,\tiny{$\pm$0.0132} & 0.6025\,\tiny{$\pm$0.0025} & 0.983\,\tiny{$\pm$0.002} & 0.600\,\tiny{$\pm$0.017} \\
\textbf{\spanbalance} & \textbf{0.7695}\,\tiny{$\pm$0.0087} & 0.6029\,\tiny{$\pm$0.0096} & 0.972\,\tiny{$\pm$0.003} & 0.624\,\tiny{$\pm$0.006} \\
\midrule
\multicolumn{5}{@{}l}{\textit{forgetting mitigation}} \\
\textsc{Replay}$_{\text{single}}$ & 0.6899\,\tiny{$\pm$0.0156} & 0.5790\,\tiny{$\pm$0.0088} & 0.797\,\tiny{$\pm$0.011} & \textbf{0.783}\,\tiny{$\pm$0.010} \\
\textsc{Replay}$_{\text{cover}}$ & 0.5706\,\tiny{$\pm$0.0142} & 0.5294\,\tiny{$\pm$0.0114} & 0.871\,\tiny{$\pm$0.005} & \textbf{0.783}\,\tiny{$\pm$0.013} \\
\bottomrule
\end{tabular}
\caption{Forgetting-mitigation controls, \hqa, \qwen. Both replay arms carry a
citation share of supervised tokens ($30.7\%$ and $30.3\%$) close to
\classbalance's $33.3\%$, so the comparison holds the budget roughly fixed
and varies only whether it is delivered by data volume or by loss weight.
} 
\label{tab:forgetting}
\end{table}

\noindent \textsc{Replay}~\cite{chaudhry2019tiny, kotha2026replaying} empirically varies the class emphasis by changing volume instead of objective weight.
We run two variants of \textsc{Replay} on HousingQA that hold that share fixed and vary only how it is spread. \textsc{Replay}$_{\text{single}}$ draws one
sample from the unmasked corpus to mix back into the masked training set, dropping the citation share of supervised tokens from $81\%$ to $30.7\%$, and trains on it for all three epochs. Thus, only $59$ of $1{,}176$ citations are ever supervised, and the same $59$ three times. \textsc{Replay}$_{\text{cover}}$ keeps the aggregate share at $30.3\%$ but 
draws a \emph{disjoint} fold each epoch, 
which increases coverage at identical intensity and identical step count. It is implemented by tripling the corpus and running a single epoch, each copy carrying its own mask, so an example's citation is supervised in at most one of its three appearances.

In Table~\ref{tab:forgetting}, we see that volume does not reproduce weighting: at a matched share \textsc{Replay}$_{\text{single}}$ reaches $0.6899$ held-in AUC against \classbalance's $0.7599$. The shortfall is not coverage: raising the number of distinct citations supervised sixfold, at identical intensity and step count, moves held-in AUC \emph{down} to $0.5706$, barely above the untuned base. The missing ingredient is rather structure, which is exactly what our group-balanced objective supplies. However, while both \textsc{Replay} arms also lag \ourmethod in terms of held-in citation recall, they hold the best held-out citation recall in the table ($0.783$, above the base's $0.752$), so we can see them achieving high preservation (but they pay for with a substantial injection cost).

Meanwhile, \noindent \textsc{WiSE-FT}~\cite{wiseft} simply creates a weighted combination of the fine-tuned and base models:
$\theta_\alpha = (1-\alpha)\,\theta_0 + \alpha\,\theta_{\text{ft}}$.  Under LoRA
$\theta_{\text{ft}} = \theta_0 + \Delta W$, so we can implement \textsc{WiSE-FT} and sweep different values for it simply by a rescaling of the
weight update that costs no training at all.

In Fig.~\ref{fig:wise}, we see that \textsc{WiSE-FT} reaches a different part of the frontier, not a better one. Scaling down an aggressive masked-SFT update sits \emph{on or below} the \textsc{Sft}$\leftrightarrow$masked-\textsc{Sft} chord on held-in recall and only modestly above it on held-out ($+0.073$ to $+0.141$), where budget correction stands two to three times further above the same line ($+0.229$, $+0.272$). Scaling down \textsc{Sft} never exceeds \textsc{Sft}'s own AUC at all, so only the masked endpoint was ever a candidate to reach the high-AUC regions.

As a specificity control (not plotted), we measure held-out AUC, which we find stays flat across the
entire sweep: $0.570$--$0.593$ for every \textsc{WiSE} arm against an untuned base of $0.541$, in contrast to held-in AUC which climbs from $0.619$ to $0.746$. The change in within-group AUC therefore measures injection rather than a general shift in discriminability. Note that \classbalance and \spanbalance reach $0.6025$ and $0.6029$, the highest in the comparison.

The takeaway is that both anti-forgetting techniques buy real held-out citation recall at the cost of knowledge injection ability. Neither \textsc{Replay} nor \textsc{WiSE-FT}, however, reaches \ourmethod's injection at a comparable citation budget. They are points on the injection-preservation Pareto frontier that trade injection away for preservation, which is the opposite of what we are asking the method to do; they remain so at matched budget, at sixfold \textsc{Replay} coverage, and across the whole \textsc{WiSE-FT} interpolation path.

\subsubsection{Memory Editing}
\label{app:memory-edit}
\begin{figure}[t]
\centering
\includegraphics[width=0.78\columnwidth]{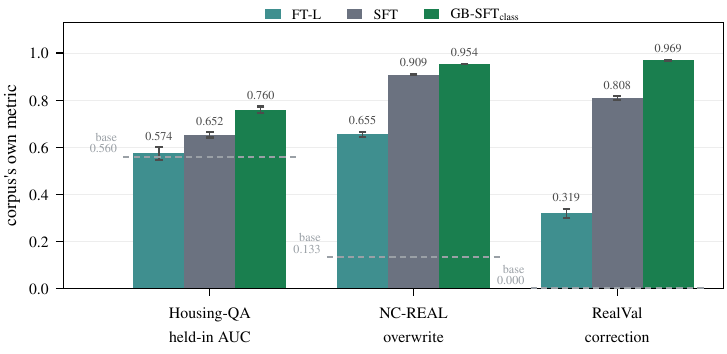}
\caption{\textbf{Single-layer editing does not recover what reweighting does.}
FT-L-style adaptation confined to layer $8$'s MLP down projection at
$5{\times}10^{-4}$ against ungrouped \sft and
\classbalance on all three legal fact-overwrite corpora.
Each corpus carries its own measure (held-in AUC, overwrite, correction)
so a bar is comparable only to the others in its group.
}
\label{fig:ftl}
\end{figure}

FT-L \citep{meng2022rome} restricts a factual edit to one layer's MLP output projection, on the view that a fact can be written to a small, identifiable part of the network. We implement its defining restriction of adaptation confined to a single layer $\ell$'s down-projection with every other adapter parameter frozen; our implementation does not reproduce its $L_\infty$ clamp on the weight delta, but captures the primary single-layer editing mechanism of FT-L.  Because a single layer trains under very different conditions from a full-network update, we swept both free settings rather than inheriting ours: layers $\{2,4,6,8,16,24\}$ of $32$ against learning rates $\{10^{-4}, 5{\times}10^{-4}, 2{\times}10^{-3}\}$, on \hqa at one seed, then carried the winning cell to all three corpora at three seeds. 

FT-L clearly learns, clearing the untuned base on all three corpora particularly its own best setting of knowledge editing (\ncreal and \realval).  However, it is still below ungrouped \sft on all three corpora and far below \classbalance (Figure~\ref{fig:ftl}). Confining the update to one layer therefore does not recover what reweighting it does.

\section{Extended Experiments and Analyses}
\label{app:extended}
\subsection{The Roles of $\alpha$ and $\beta$}
\label{app:alpha-beta}
The two exponents of Eq.~\ref{eq:gb-sft} are not symmetric in role, and we do
not offer them as a two-parameter search space. $\alpha$ is the dial, and $\beta$ is
a structural choice of \emph{which unit carries a budget}, which we exercise
only at its endpoints.

Our results provide the measurements that justify this asymmetric framing. App.~\ref{app:dose-response} shows that
the class budget $\alpha$ induces accounts for held-in AUC to within seed noise
($r = -0.945$), and that $\beta$ acts only through that budget: at $\alpha = 0$,
where the class split is pinned at $1/\numClasses$ whatever $\beta$ does, the
two endpoints (pooling spans and pooling tokens, structurally different
operations) land within $0.0002$ of each other on \qwen\ and within $0.0029$
on \llama. We therefore report one tunable dial and one structural switch, and
recommend treating $\beta$ as binary.

Recovering \sft, \classbalance and \spanbalance as corners of one object keeps
the comparison internal to a single objective rather than a bake-off between
three. We note that this is a sanity property of the parameterization rather
than the contribution itself. Choosing a normalization unit for a sequence loss
is long-standing practice: it is a configuration flag in standard training
toolkits, and the token- versus sequence-mean choice has been studied directly
in RL post-training \citep{liu2025drgrpo}. What is new here is the class axis, a
budget over the \emph{semantic roles} of tokens within a single target sequence,
which partitions derived from sequence or sampler boundaries cannot express.

\subsection{Grouping Granularity and the Induced Budget}
\label{app:granularity}

Equal-budget-per-class is not invariant to subdividing a class. At $\alpha = 0$ each of $K$ classes receives $1/K$ of the budget regardless of class masses, so splitting one concept into two raises that concept's share. On \hqa, going from the three-class partition (answer / citation / prose) to a four-class one that splits the citation into identifier and reporter moves the citation from $.333$ to $.500$ of the budget. This is a change the dose-response of Fig.~\ref{fig:dose-response} predicts should cost accuracy before the arm is run.

Empirical results confirm this prediction. Table~\ref{tab:granularity} orders every \hqa grouping we have by the citation share it induces; held-in AUC falls monotonically as that share rises, whether the rise comes from subdividing a class or from moving $\alpha$ within a fixed partition.

\begin{table}[h]
\centering
\small
\begin{tabular}{lccc}
\toprule
Grouping & answer share & citation share & held-in AUC \\
\midrule
3-class, $\alpha = 0$ & $.333$ & $.333$ & $0.7721$\,\tiny{$\pm$0.0136} \\
3-class, $\alpha = .2072$ & $.250$ & $.444$ & $0.7559$\,\tiny{$\pm$0.0080} \\
4-class, $\alpha = 0$ & $.250$ & $.500$ & $0.7536$\,\tiny{$\pm$0.0054} \\
3-class, $\alpha = .3084$ & $.213$ & $.500$ & $0.7414$\,\tiny{$\pm$0.0069} \\
\bottomrule
\end{tabular}
\caption{\textbf{Granularity acts through the budget it induces.} \hqa, \qwen, three seeds per cell. The four-class arm loses $0.0185$ to the three-class arm, which the dose-response predicts from the citation share alone. At matched citation share ($.500$) the four-class and three-class arms differ by $0.0122$, which the residual $.037$ answer-share gap accounts for.}
\label{tab:granularity}
\end{table}

We do note that this control has a limitation: the three-class partition has one free parameter while the four-class target has two degrees of freedom, so each matched control fixes one coordinate of the budget and leaks the other. Matching on citation share, the coordinate Fig.~\ref{fig:dose-response} identifies as active ($r = -0.945$), leaves an answer-share gap that favors the four-class arm; matching on answer share instead leaves a citation gap large enough to account for most of the effect under test. The two cannot be matched simultaneously inside this family, and that is a property of the $\alpha = 0$ corner rather than of these runs.

\subsection{Composing with Per-Token Scoring}
\label{app:composition}

Our weights alter the loss's \emph{aggregation geometry}; Rho-1, DFT and related methods alter its \emph{per-token scoring function}. The two are compatible, since $\omega_t$ multiplied by any per-token statistic is well defined. To offer insights into how these two classes of modifications can compose, we report the results of one such composition. Placing DFT’s per-token weights inside our groups on \hqa raises held-in AUC to $0.7901 \pm 0.0171$ from $0.7721 \pm 0.0136$, while held-out AUC falls from $0.6043 \pm 0.0225$ to $0.5422 \pm 0.0100$. This composition behaves like the rest of the family: it buys acquisition with preservation, which is the exchange Fig.~\ref{fig:dose-response} already describes rather than an escape from it.

Read across the three paired seeds, the two sides of the exchange uphold our default recommendation of uniform weighting within spans. The held-in gain is $+0.018$ ($t(2) = 1.01$, $p = 0.42$), and one of the three seeds moves against it; the held-out loss is $-0.062$ and consistent in all three ($t(2) = -3.61$, $p = 0.069$). We therefore report the composition as evidence that the two axes are compatible rather than as a recommendation: at this sample size its benefit sits inside seed noise and its cost does not. The deployment setting where it could be worth revisiting is one where held-out preservation carries no weight, such as a closed corpus serving only facts it was trained on.

\subsection{Loss-Based Group Reweighting as a Baseline}
\label{app:ye-baseline}

App.~\ref{app:related} distinguishes our weighting from \citet{ye2025disentangling} structurally. Here we run their objectives under our own grouper, so that only the weighting rule differs.

Their fixed-weight variant, $\alpha$-FT, coincides with our $\beta = 1$ family when its coefficient equals a corpus's payload share. At that matched coefficient the two are not distinguishable: on \hqa, $0.7792 \pm 0.0232$ against our $0.7721 \pm 0.0136$; on \ncreal, held-in overwrite $0.9544 \pm 0.0049$ against our $0.9497 \pm 0.0053$. We read the coincidence as evidence for the parameterization rather than against it. Equal budget is what $\alpha = 0$ \emph{means} in our family, on any corpus, with no tuning, whereas reaching it in theirs requires knowing each corpus's composition and setting the coefficient accordingly. Our corpora span payload shares from $14.9\%$ to $0.89\%$, so the coefficient that reaches it differs for every one.

We find that the new baselines behave like existing baselines of ours. RFT reaches $0.9053 \pm 0.0035$ on \ncreal, in line with ungrouped \sft rather than with budget correction. And $\alpha$-FT in its $\alpha \to 0$ limit suppresses the boilerplate class almost entirely and reaches $0.9702 \pm 0.0025$, above our $\alpha = 0$ arm on the injection axis. That limit is masking by another name, and the trade it makes is the one Table~\ref{tab:hqa-both} already reports for \maskedsft: injection bought at the cost of scaffold preservation. We note it because the comparison is only informative when both axes are read together.

\subsection{Joint Answer-and-Citation Correctness under Free Generation}
\label{app:jointgen}

The paper reports ranking and free generation separately. Here we report the conjunction of answer and citation correctness under free generation, a stricter measurement. 

On \hqa held-in (one seed, $150$ items), \ourmethod leads on both components, answer $0.813$ against $0.780$ and exact citation $0.407$ against $0.360$, while the joint rate is $0.493$ against $0.487$. The component gains do not compound, which suggests the two error types are anti-correlated rather than independent. Held-out joint correctness is near zero for every arm, as expected, since held-out citations are not taught.

We think the gap is knowledge-limited rather than decoding-limited, and the evidence is that constraining decoding does not close it: forcing a well-formed citation scores $0.560$ against $0.580$ for ordinary greedy decoding. If the model had learned the payload and were failing only to express it, constraining the surface form would help. It does not. That points to the corpus-level remedy \citet{allenzhu2024physics} identify: a fact becomes extractable once it appears under sufficient paraphrase and format diversity. This remedy is orthogonal to how the loss is weighted, and we consider it the most promising direction for future work that this measurement opens.

\subsection{\sftgo with Oracle Grouping, Across Benchmarks}
\label{app:sftgo-grid}

Fig.~\ref{fig:tfidf_vs_oracle} shows that good grouping is a large share of the
effect for both objectives. The complementary question may be asked: given
the \emph{same} oracle grouping, does the objective still matter? Table~\ref{tab:sftgo-grid}
collects every corpus and model where we ran both, on each benchmark's primary
injection metric, and separately on the preservation metric where \hqa\ provides one.

\begin{table}[h]
\centering
\small
\begin{tabular}{llcccc}
\toprule
Corpus & Model & Metric & \classbalance & \spanbalance & \sftgo$_{\text{oracle}}$ \\
\midrule
\multicolumn{6}{@{}l}{\textit{injection axis}} \\
\hqa     & \qwen  & AUC-in    & .760 & \best{.769} & .756 \\
\hqa     & \llama & AUC-in    & .748 & .765 & \best{.773} \\
\ncreal  & \qwen  & ovwrite   & \best{.954} & \best{.954} & .910 \\
\ncreal  & \llama & ovwrite   & \best{.947} & .946 & .934 \\
\openfda & \qwen  & ovwrite   & \best{.799} & .788 & .605 \\
\openfda & \llama & ovwrite   & .842 & \best{.852} & .747 \\
\midrule
\multicolumn{6}{@{}l}{\textit{preservation axis}} \\
\hqa     & \qwen  & cite-out  & .600 & .624 & \best{.720} \\
\hqa     & \llama & cite-out  & .669 & .723 & \best{.815} \\
\bottomrule
\end{tabular}
\caption{\textbf{The objective is not inert given the grouping, and the ordering
depends on the axis.} Both methods receive the same oracle grouping; only the
weighting rule differs. \ourmethod leads on injection in five of six cells, by
margins up to $.194$; \sftgo$_{\text{oracle}}$ leads the sixth (\hqa/\llama
AUC-in, by $.008$) and leads both preservation cells. On \realval, \sftgo$_{\text{oracle}}$
falls below ungrouped \sft\ for \qwen (Fig.~\ref{fig:correction}).}
\label{tab:sftgo-grid}
\end{table}

The split is not arbitrary. \sftgo\ optimizes a mixture of worst-group and
average loss, so it adaptively emphasizes whichever group is currently hardest.
On fact-editing corpora that group is the long scaffold the model has least
reason to fit. This is why the same mechanism that costs it injection buys
it scaffold preservation, and why the ordering between the two methods flips
with the axis being measured. Our budget is fixed by the grouping rather than
chased during training, so it does not inherit that behavior.

\subsection{Additional Related Work}
\label{app:related}
\citet{ye2025disentangling} share our premise that tokens serving different roles should not be weighted equally in \sft. They identify two groups with a shuffle-aware discriminator and reweight the per-token cross-entropy accordingly. Because the two objectives are close in motivation, it is worth being precise about where they differ and where they agree.

\paragraph{Group weighing differences.}
\citet{ye2025disentangling} 's group weight $\omega_g \propto \exp(\bar{L}_g/\tau)$ depends on each group's \emph{current mean loss} and multiplies that group's \emph{summed} loss, so a group's gradient mass remains proportional to its token count. The $1/\tau \to 0$ limit recovers standard \sft; the $\tau \to 0$ limit approaches suppressing the boilerplate entirely. Equal loss budget is a limit of that family rather than a point in it.

Our weight depends instead on group \emph{cardinality}, $m_\aclass$ and $n_\aspan$. That is what makes $\alpha = 0$ and $\beta = 0$ reachable corners: equal budget is what $\alpha = 0$ \emph{means}, on any corpus, with no tuning. Their split is also binary and span-agnostic, with no refinement level.

\paragraph{Recovering \classbalance under a special case.}
Their fixed-weight variant, $\alpha$-FT, coincides with our $\beta = 1$ family when its coefficient is set to a corpus's payload share. We ran $\alpha$-FT and RFT under our own grouper and confirm their agreement at the matched coefficient: the objectives are the same up to numerical precision, and the held-in difference between them is not statistically distinguishable.

We read the coincidence as evidence for the parameterization rather than against it. Equal budget is a distinguished point of a family that $\alpha$-FT reaches only when its free coefficient is tuned to each corpus's composition; our corpora span payload shares from $14.9\%$ to $0.89\%$, so the coefficient that reaches it differs for every one. Expressing the weight as a function of cardinality makes that point addressable by construction rather than by search.

\paragraph{Weighting across sequences.}
\citet{hsieh2023distilling} meet the same pathology from the other side. Their $\lambda$ weights two \emph{separate} output sequences, a label and a rationale, selected by \texttt{[label]} and \texttt{[rationale]} prefixes and trained as distinct tasks; the division is therefore made before the loss rather than inside it, and the paper neither reports a value for $\lambda$ nor ablates it. The construction is a multi-task decomposition, not a token-level reweighting. It does not apply to a corpus whose payload and scaffold arrive in one continuation, which is the setting we address and the one their own Eq.~5 sets aside.

That ablation is nonetheless the closest independent evidence for our premise that we are aware of. Concatenating rationale and label into a single sequence under uniform cross-entropy can score \emph{below} ordinary finetuning: adding supervision hurts, because the added tokens take budget from the ones that carry the answer. They remedy it by splitting the sequence; we remedy it by reallocating the budget within the sequence, which is available when the split is not. We read the two as complementary responses to one failure mode rather than as competing methods.

\end{document}